\documentclass[12pt]{article}
\usepackage{amssymb,amsmath,amsfonts,eurosym,ulem,graphicx,
	    caption,color,setspace,comment,footmisc,
	      pdflscape,subcaption,array, multicol,multirow,tikz, bm,
            booktabs, rotating, titling}
\usepackage[utf8]{inputenc}
\usepackage[left=1.0in,right=1.0in,top=1in,bottom=1in]{geometry}
\usepackage{indentfirst}
\usepackage[sort]{natbib}
\usepackage[colorlinks=true, allcolors=blue]{hyperref}
\newcommand{\doi}[1]{\url{#1}}
\usetikzlibrary{arrows,shapes,positioning,intersections,calc}
\newcolumntype{L}[1]{>{\raggedright\arraybackslash}p{#1}}
\newcolumntype{C}[1]{>{\centering\arraybackslash}p{#1}}
\newcolumntype{R}[1]{>{\raggedleft\arraybackslash}p{#1}}
\makeatother

\usepackage{tocloft}
\usepackage{setspace} 
\usepackage{titletoc}

\usepackage{caption}
\usepackage{adjustbox}
\makeatletter
\renewcommand*\env@matrix[1][\arraystretch]{%
  \edef\arraystretch{#1}%
  \hskip -\arraycolsep
  \let\@ifnextchar\new@ifnextchar
  \array{*\c@MaxMatrixCols c}}
\makeatother

\newcommand{\subtitle}[1]{\posttitle{\par\end{center}\begin{center}\large#1\end{center}\vskip0.5em}}

\title{HMM-LSTM Fusion Model for Economic Forecasting}

\author{Guhan Sivakumar}
\date{May 8, 2024}

\begin{document}

\maketitle
\begin{abstract}
    This paper explores the application of Hidden Markov Models (HMM) and Long Short-Term Memory (LSTM) neural networks for economic forecasting, focusing on predicting CPI inflation rates. The study explores a new approach that integrates HMM-derived hidden states and means as additional features for LSTM modeling, aiming to enhance the interpretability and predictive performance of the models. The research begins with data collection and preprocessing, followed by the implementation of the HMM to identify hidden states representing distinct economic conditions. Subsequently, LSTM models are trained using the original and augmented data sets, allowing for comparative analysis and evaluation. The results demonstrate that incorporating HMM-derived data improves the predictive accuracy of LSTM models, particularly in capturing complex temporal patterns and mitigating the impact of volatile economic conditions. Additionally, the paper discusses the implementation of Integrated Gradients for model interpretability and provides insights into the economic dynamics reflected in the forecasting outcomes.
\end{abstract}
\doublespacing

\thispagestyle{empty}

\newpage
\setstretch{1.75}
\setcounter{page}{1}
\begin{spacing}{1.39}
    {
      \hypersetup{linkcolor=black}
      \tableofcontents
    }
\end{spacing}

\newpage
\section{Introduction}
Forecasting CPI Inflation has been a challenge for several years due to the complexity of economic dynamics. In recent years, advanced time series machine learning models have been applied to solve economic problems, and they have been shown to perform better than traditional forecasting methods; however, the major drawback is the interpretability of the models which is crucial for economic analysis. This paper explores the integration of Hidden Markov Models (HMM) and Long Short-Term Memory (LSTM) neural networks to forecast CPI Inflation, as well as implements Integrated Gradients for feature explainability.

Major economic indicators were selected as features, and the selected economic indicators provide valuable insights into different aspects of the economy, ranging from monetary policy to market sentiment. The forecasting model gains a comprehensive understanding of the various factors influencing inflation rates and mitigates the risk of relying on just one variable. With multiple features, the model can learn and identify different economic trends and adjust the forecasts accordingly. Appropriate data cleaning was done to handle the different variables and ensure consistency throughout the modeling process.

An HMM analysis was first done to determine the validity of the features vs National Bureau of Economic Research (NBER) recession indicators, and subsequently the hidden states and means derived from the HMM were used as additional features in the LSTM network. LSTM networks work similarly to the human brain and perform significantly better than artificial neural networks when dealing with time-dependent data. By augmenting the features, the LSTM is given more information regarding structural trends present in the data. Different feature combinations were tested on the LSTM and validated thoroughly to ensure optimal performance. Finally, a similar model architecture was then used to get 1-year and 2.5-year forecasts and compared against official forecasts. 

To overcome the interpretability issue, traditional methods such as Shapley Values and Local Interpretable Model-Agnostic Explanations (LIME) were tested; however, they have not been adapted yet to work with all custom models. So, a less commonly used technique in time series forecasting, Integrated Gradients, was used to get an approximation of feature importance. Though not perfect, it can provide insights into the LSTM network which is generally considered a ``Black-Box" model. Identifying the important economic features gives insights into how the different features affect the forecast, which can be crucial to economists and stakeholders. 

This research aims to solve some of the major issues of economic forecasting with advanced machine learning techniques. By creating a fusion model, the forecasts are more robust and the addition of integrated gradients provides interpretability needed for economic analysis. Bridging the gap between traditional economic modeling and advanced machine learning offers a promising avenue for more robust and insightful economic forecasts. As technology continues to evolve, the integration of innovative methodologies becomes essential for policymakers, analysts, and researchers.

\newpage
\section{Literature Review}
\subsection{Machine Learning Methods in Economics}
Athey and Imbens' paper discusses the effectiveness of machine learning methods and their application in economics. Specifically, ``[neural networks] have been found to be very successful in complex settings, with extremely large number of features" \citep{paper1}. The addition of multiple hidden layers allows the models to capture non-linear trends between the features, and they mention that the main interest is in ``the properties of the predictions from these specifications" \citep{paper1}, and not the parameter estimates. Traditional methods to solve for the non-linear weights would lead to poor results, so different algorithms have been implemented, such as back-propagation. The growing success of machine learning methods in economics, highlighted by the effectiveness of neural networks in capturing complex relationships, signifies their potential to enhance forecasting.  

\subsection{Forecasting China's GDP Using LSTM and HMM}
In this paper, the authors discuss the advantages of using an LSTM-HMM approach to forecast GDP fluctuation states for China's GDP. Compared to a standard HMM, ``[they] select the predicted real time CPI fluctuation state by LSTM to be the observable state when we forecast the real time GDP fluctuation state, instead of the observed CPI fluctuation states with time lag" \citep{paper2}. This approach led to better performance and more consistent results in their long-term window testing, most notably ``within the 10-year time window, ..., the LSTM-HMM monthly univariate input has the best performance in prediction" \citep{paper2}. The integration of LSTM and HMMs in forecasting China's GDP offers improved accuracy and stability, particularly evident in long-term predictions, showcasing the efficacy of this combined approach for economic forecasting.

\subsection{Deep Learning Applications in Economic Forecasting}
With the rise in deep learning, the authors explore various architectures and determine which models are preferred and successful in economics. Since RNNs only consider the most recent state, ``the cellular state of LSTM determines which states should be left behind and which states should be forgotten. Hence, LSTM plays an important role in many fields of economic research" \citep{paper3}. In a list of papers that utilize RNNs. specifically LSTMs, they noticed that the LSTM model outperformed traditional econometric approaches such as ARIMA, MA, vanilla neural networks, etc. Some of the drawbacks they mention for RNNs are, ``there are many parameters that need to be trained, which are prone to gradient dissipation or gradient explosion; [and they are] without feature learning ability" \citep{paper3}. The authors highlight the superiority of LSTMs over traditional econometric methods, despite noted challenges like parameter tuning and feature explainability.

\subsection{Economic State Prediction Using Hidden Markov Models}
The author explores the effectiveness of HMMs to predict the economic state of various countries such as the US, UK, Japan, etc. Various features relevant to the economy were selected to further enhance the capability of these models. Rather than just testing 2-state models, the paper explores using more states, as well as various transformations on the series, to determine if the performance of HMMs improves. From the results, ``the two-state, and especially the three-state higher-order hidden Markov model can predict economic states. The three-state model outperforms all viable benchmarks and other HMMs" \citep{paper4}. Though the 3-state model performs well, the downside is that it is not easily classifiable as the 2-state model, so meaningful results become harder to derive. Despite these drawbacks, HMMs can still be used to identify new economic periods and provide information that may not be captured in traditional classification models. 

\subsection{Interpretable Neural Networks for Panel Data Analysis}
Despite the popularity of advanced machine learning methods, ``Economists are still nervous about using more advanced tools like neural networks, since they mostly deliver outcomes from a complicated black box without transparency or interpretability" \citep{paper5}. To solve this issue, the authors created a modified neural network that achieves both high accuracy and interpretability. Their results demonstrate the power of an interpretable neural network, and that ``the increase in interpretability may allow models to achieve higher accuracy in the test set than unrestricted models" \citep{paper5}. This paper highlights a new approach to the importance of transparency in black-box models, and the rise in Explainable AI has led to model agnostic interpretability, such as Shapley values, a concept first introduced to solve a problem in cooperative game theory. 

\subsection{Forecasting Economics and Financial Time Series with LSTM}
To emphasize the viability of machine learning methods for forecasting, this paper focuses on comparing the traditional ARIMA model to an LSTM network. The two models were tested on financial data, such as IXIC, GSPC, DJI, etc., and economic data such as housing, medical care, exchange rate, etc., and ``the study shows that LSTM outperforms ARIMA. The average reduction in error rates obtained by LSTM is between 84 - 87 percent when compared to ARIMA indicating the superiority of LSTM to ARIMA" \citep{paper6}. The paper also discusses the importance of hyper-parameter tuning and overfitting with regards to the LSTM, and they noticed that there was ``no improvement when the number of epochs is changed" \citep{paper6}. The results from this paper demonstrate the predictive power of LSTMs and their application in finance and economics. Advancements in deep learning will eventually lead to better models and even outperform LSTMs.

\newpage
\section{Data Collection and Cleaning}
\subsection{Feature Engineering}
\begin{figure}[!ht]
    \centering
    \includegraphics[width=0.95\textwidth]{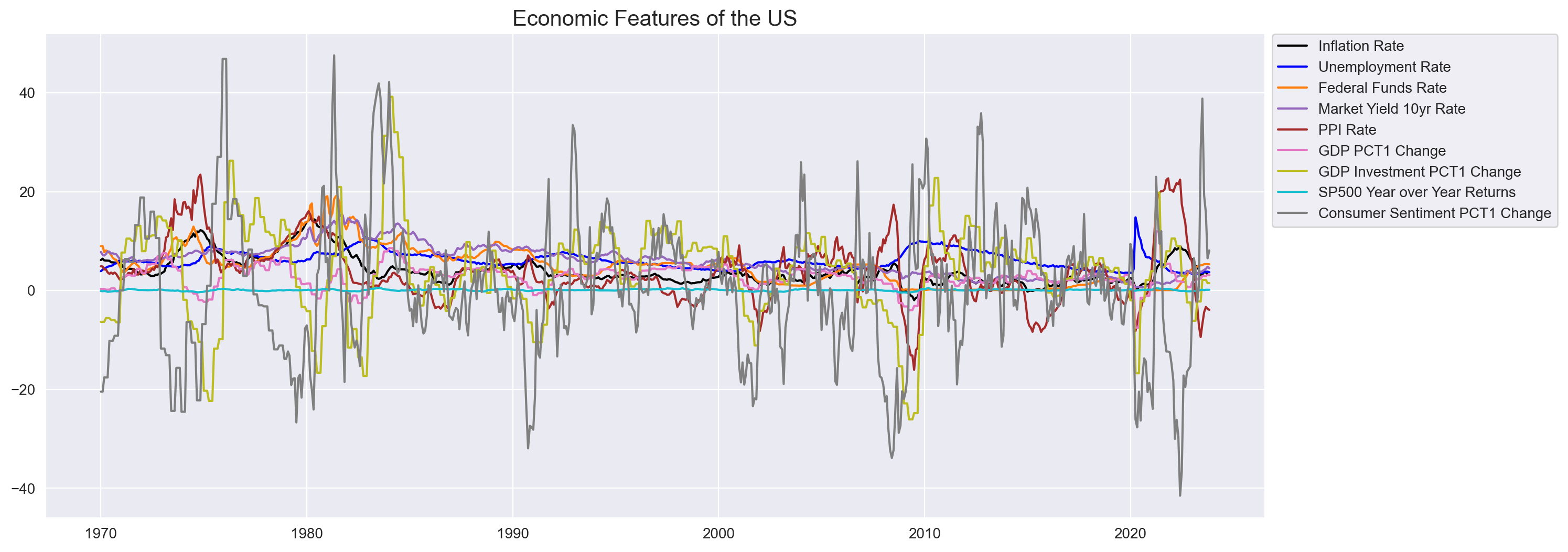}
    \caption{Economic Features of the US}
    \label{fig:econfeat}
\end{figure}

The data was collected from \citep{fred} and \citep{yfinance}. To cover various aspects of the economy, several key features were selected, such as Federal Funds Rate, Unemployment Rate, GDP growth rate, PPI Rate, etc. Since some of the data was quarterly and daily, interpolation and aggregation were done respectively to make the data into monthly time steps. The features were converted to growth rates to ensure consistency with the target variable, the CPI Inflation Rate. The data ranges from 1970-01-01 to 2023-01-01, and \autoref{fig:econfeat} is a plot that provides an overview of the features.
 
Descriptive statistics are provided in \autoref{tab:summarytable}. There are a total of 647 monthly observations, and the table gives us historical information about the economy. 
\begin{table}[!ht]
    \centering
    \begin{adjustbox}{width=0.9\textwidth}
    \begin{tabular}{lrrrrrrrrrrr}
\toprule
 & inflation\_rate & unemployment\_rate & federal\_funds\_rate & market\_yield\_10\_rate & ppi\_rate & gdp\_pc1 & gdp\_invest\_pc1 & sp500\_yoy & consumer\_senti\_pc1 & oil\_price\_pc1 & recessions \\
\midrule
count & 647.0000 & 647.0000 & 647.0000 & 647.0000 & 647.0000 & 647.0000 & 647.0000 & 647.0000 & 647.0000 & 647.0000 & 647.0000 \\
mean & 4.0399 & 6.1306 & 4.9066 & 6.0302 & 3.9061 & 2.7262 & 3.9420 & 0.0853 & 0.3952 & 12.4316 & 0.1314 \\
std & 2.9384 & 1.7038 & 3.9166 & 3.1398 & 6.0956 & 2.3174 & 9.5545 & 0.1616 & 13.8638 & 41.6073 & 0.3381 \\
min & -1.9588 & 3.4000 & 0.0500 & 0.6200 & -16.0584 & -7.5285 & -26.0881 & -0.4251 & -41.5205 & -74.0839 & 0.0000 \\
25\% & 2.1086 & 4.9000 & 1.4250 & 3.5450 & 0.4937 & 1.6650 & -1.4419 & -0.0057 & -7.3773 & -8.8633 & 0.0000 \\
50\% & 3.1940 & 5.8000 & 4.9900 & 5.9800 & 3.5683 & 2.8589 & 4.7050 & 0.1015 & 0.0000 & 5.1300 & 0.0000 \\
75\% & 5.0354 & 7.2000 & 6.9100 & 7.8850 & 6.3669 & 4.1878 & 9.6544 & 0.1922 & 7.3677 & 24.4842 & 0.0000 \\
max & 14.5923 & 14.8000 & 19.1000 & 15.3200 & 23.4409 & 11.9503 & 39.1907 & 0.5264 & 47.5822 & 272.9305 & 1.0000 \\
\bottomrule
\end{tabular}

    \end{adjustbox}
    \caption{Summary Statistics}
    \label{tab:summarytable}
\end{table} 

\noindent Without accounting for breaks, the average inflation rate has been around 4\%, and the current inflation rate as of March 2024 is 3.475\%. Unemployment rate is below its historical average, and due to the dual mandate described in \citep{dualmandate}, there has not been any pressure to decrease interest rates from 5.33\%. West Texas Intermediate (WTI) Crude Oil price was considered as a feature, but due to its range and volatility, it was not included in the model. Information regarding oil price and its effects can be found in \autoref{section:oilprice}.

\subsection{Correlations and Stationarity}
\begin{figure}[!ht]
    \centering
    \includegraphics[width=0.53\textwidth]{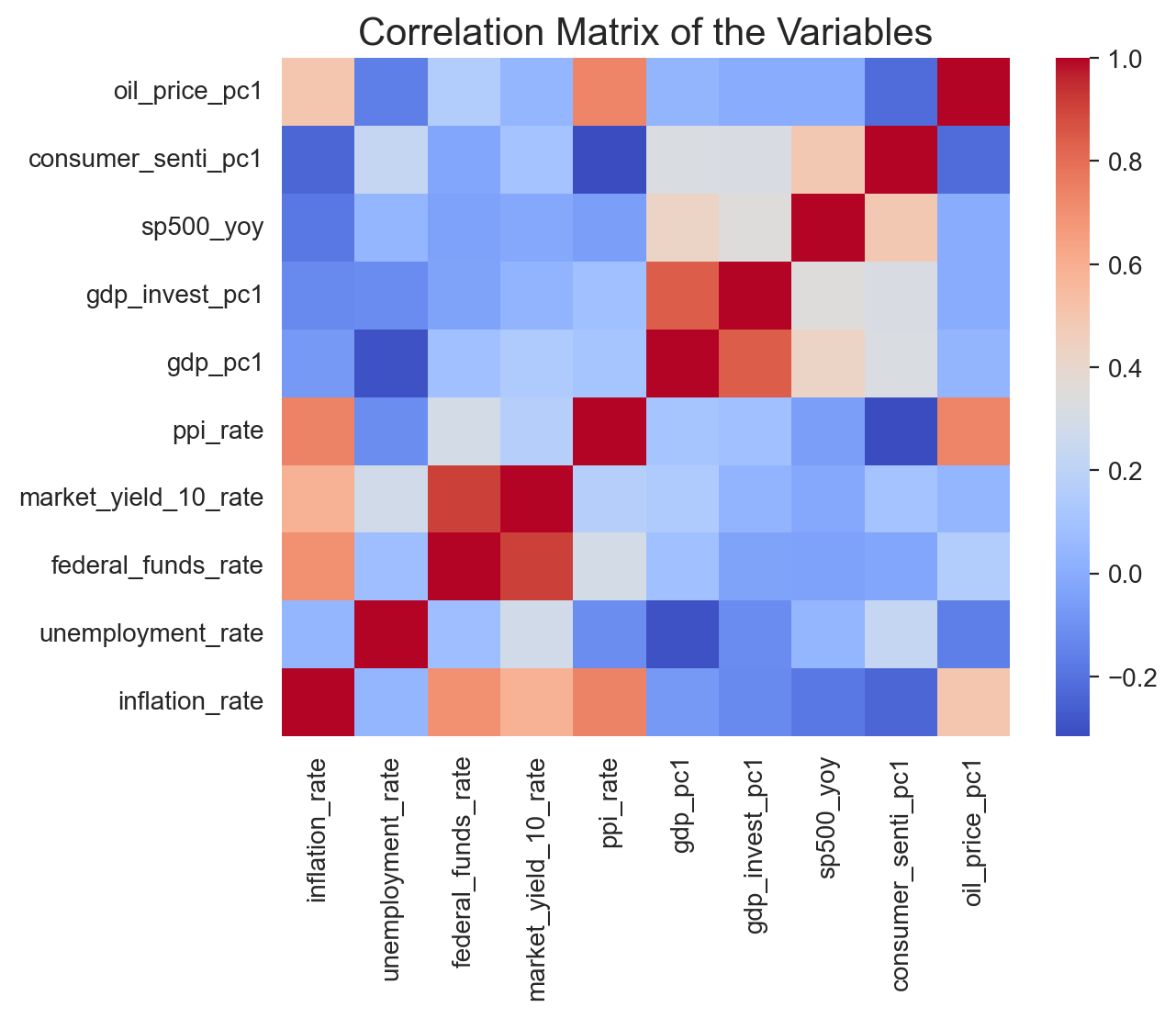}
    \caption{Correlation Matrix of the Variables}
    \label{fig:corrmat}
\end{figure}

The correlation matrix in \autoref{fig:corrmat}, shows that there are no unexpected significant correlations with inflation rate. While multicollinearity exists between certain variables like GDP and GDP Investment, each contributes differently to inflation forecasting. GDP Investment acts as a leading indicator for predicting inflation, and GDP offers insights into overall economic growth. 

The Augmented Dickey-Fuller (ADF) test by \citep{adftest} was conducted to test stationarity and the results are found in \autoref{tab:adftest}. Only two of the series were non-stationary, but HMMs and LSTMs are able to capture non-linear relationships making them more robust to non-stationary trends.

\newpage
\section{Hidden Markov Models}
HMMs offer a powerful framework, created by \citep{hmm}, for capturing the dynamic nature of economic systems, and they work by determining latent or unobservable states in time series data. It is a probabilistic model and aims to maximize the likelihood of being in a particular state at a given time step. For this paper, four hidden states were used, and the states are indicative of the prevailing economic environment during those periods. There are three fundamental steps to HMMs at each time step, determine initial probabilities, compute the probability of the states, decode the probabilities and determine the latent state.

\subsection{Expectation Maximization and Viterbi Decoding}
The Expectation Maximization (EM) algorithm is a numerical method used to estimate the Maximum Likelihood Estimator. There are three main probabilities that the HMM computes, initial, transmission, and emission probabilities; however, the implemented method for this paper assumes Gaussian emissions. The algorithm requires prior knowledge, so the Dirichlet distribution, a common distribution used as a prior for Bayesian estimation, is used as the prior. The EM algorithm by \citep{emalgorithm}, is described in \autoref{eq:emalgo}
\begin{align}\label{eq:emalgo}
    Q({\boldsymbol {\theta }}\mid {\boldsymbol {\theta }}^{(t)})&=\operatorname {E} _{\mathbf {Z} \sim p(\cdot |\mathbf {X} ,{\boldsymbol {\theta }}^{(t)})}\left[\log p(\mathbf {X} ,\mathbf {Z} |{\boldsymbol {\theta }})\right]\\
    {\boldsymbol {\theta }}^{(t+1)}&={\underset {\boldsymbol {\theta }}{\text{argmax}}}\ Q({\boldsymbol {\theta }}\mid {\boldsymbol {\theta }}^{(t)})\notag\\
    \implies {\boldsymbol {\theta }}^{(t+1)}&={\underset {\boldsymbol {\theta }}{\text{argmax}}}\operatorname {E} _{\mathbf {Z} \sim p(\cdot |\mathbf {X} ,{\boldsymbol {\theta }}^{(t)})}\left[\log p(\mathbf {X} ,\mathbf {Z} |{\boldsymbol {\theta }})\right]\notag
\end{align}
where the first equation is the Expectation (E) step and the second is the Maximization (M) step. During the E-step, the algorithm computes the expected values of the latent variables given the observed data and the current parameter estimates. In the M-step, it maximizes the likelihood of the observed data under the model, updating the parameters based on the computed expected values. This iterative process continues until convergence, yielding parameter estimates that optimize the fit between the model and the observed data. 

After calculating the probabilities at each step, the Viterbi Decoding Algorithm is utilized to determine the state at the given time step. The algorithm, created by \citep{viterbialgorithm}, is given below in \autoref{eq:viterbi}
\begin{align}\label{eq:viterbi}
    \max _{k} V^{k}_{T}&= \max_{\{S_{t}\}^{T}_{t=1}}p \left(\left\{S_{t}\right\}^{T}_{t=1}, \left\{O_{t}\right\}^{T}_{t=1} \right)\\
    V_{t}^{k} &= \max_{S_{1}, ..., S_{t-1}} p(S_{t}=k, S_{1}, ..., S_{t-1}, O_{1}, ..., O_{t})\notag\\
    &= p(O_{t} \mid S_{t}=k) \max_{i} p(S_{t}=k \mid S_{t-1} = i)V^{i}_{t-1}\notag
\end{align}
where the first equation computes the maximum probability $V_{t}^{k}$ of the final state $k$ over all the possible state sequences $\{S_{t}\}^{T}_{t=1}$ given the observed sequence $\{O_{t}\}^{T}_{t=1}$. The second equation computes the maximum probability $V_{t}^{k}$ of being in state $k$ at time step $t$. It recursively calculates this probability based on the observation $O_{t}$ given state $k$ and the maximum probability of transitioning from state $k$ from the previous time step $t-1$. The algorithm provides an optimal approach to determine the most likely sequence of the hidden states based on the observed data and transition probabilities.

\subsection{Markov Chain Probabilities}
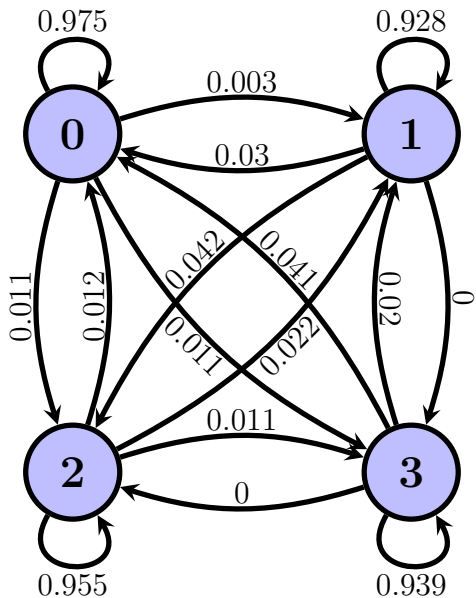
\begin{figure}[!ht]
    \hspace{4.9cm}
    \begin{tikzpicture}[scale=0.9, >=stealth]
    \tikzstyle{state}=[circle, minimum size=1.25cm, draw=black, fill=blue!25, text=black, align=center, line width=2.0pt]
    \node[state] (0) at (0,0) {\textbf{\Large{0}}};
    \node[state] (1) at (5, 0) {\textbf{\Large{1}}};
    \node[state] (2) at (0, -5) {\textbf{\Large{2}}};
    \node[state] (3) at (5, -5) {\textbf{\Large{3}}};
    
    \path[draw, line width=2.0pt]
    (0) edge[loop above, out=120, in=60, looseness=4, ->] node[yshift=-0.5mm] {$0.975$} (0)
    (0) edge[bend left=17.5, ->] node[midway, above, yshift=-1mm] {$0.003$} (1)
    (0) edge[bend right=17.5, ->] node[midway, left, xshift=-6mm, yshift=-2.1mm, sloped, rotate=180] {$0.011$} (2)
    (0) edge[bend right=17.5, ->] node[midway, left, xshift=6mm, yshift=-2.0mm, sloped] {$0.011$} (3)
    
    (1) edge[bend left=17.5, ->] node[midway, above, yshift=-1mm] {$0.03$} (0)
    (1) edge[loop above, out=120, in=60, looseness=4, ->] node[yshift=-0.5mm] {$0.928$} (1)
    (1) edge[bend right=17.5, ->] node[midway, left, xshift=6mm, yshift=2.0mm, sloped] {$0.042$} (2)
    (1) edge[bend left=17.5, ->] node[midway, right, xshift=-3.5mm, yshift=2.1mm, sloped] {$0$} (3)
    
    (2) edge[bend right=17.5, ->] node[midway, left, xshift=6mm, yshift=2mm, sloped] {$0.012$} (0)
    (2) edge[bend right=17.5, ->] node[midway, right, xshift=-6mm, yshift=-2.0mm, sloped] {$0.022$} (1)
    (2) edge[loop below, out=240, in=300, looseness=4, ->] node[yshift=0.5mm] {$0.955$} (2)
    (2) edge[bend left=17.5, ->] node[midway, above, yshift=-1mm] {$0.011$} (3)
    
    (3) edge[bend right=17.5, ->] node[midway, right, xshift=-6mm, yshift=2.0mm, sloped] {$0.041$} (0)
    (3) edge[bend left=17.5, ->] node[midway, right, xshift=6mm, yshift=-2mm, sloped, rotate=180] {$0.02$} (1)
    (3) edge[bend left=17.5, ->] node[midway, above, yshift=-1mm] {$0$} (2)
    (3) edge[loop below, out=240, in=300, looseness=4, ->] node[yshift=0.5mm] {$0.939$} (3);
\end{tikzpicture}
    \caption{Transmission Probabilities of HMM}
    \label{fig:transmats}
\end{figure}

To visualize the transmission matrix generated by the HMM, a Markov chain is provided below in \autoref{fig:transmats}. For this paper, four hidden states were used as there are multiple stages of economic recovery and expansion, and two states may not fully capture that. Though there are some probabilities that are 0, there are no absorbed states which means that the states can freely transition from one another and not get stuck in a specific state. The stationary distribution, or equilibrium distribution, of the Markov chain is as follows, State 0: 0.487, State 1: 0.133, State 2: 0.247, State 3: 0.133.

For numerical stability, a diagonal covariance was chosen since the correlation matrix determined that there were no major correlations between the target and the features. A diagonal covariance matrix assumes that there are no relationships between the different variables, whereas a full covariance matrix assumes that all the features are interrelated. Although economic theory suggests that the variables are related, the data shows weak relationships between them, and using a full covariance matrix led to absorbed states and overfitting of the model.

\subsection{Determining the Hidden States and Intuition}
\begin{figure}[!ht]
    \centering
    \includegraphics[width=0.94\textwidth]{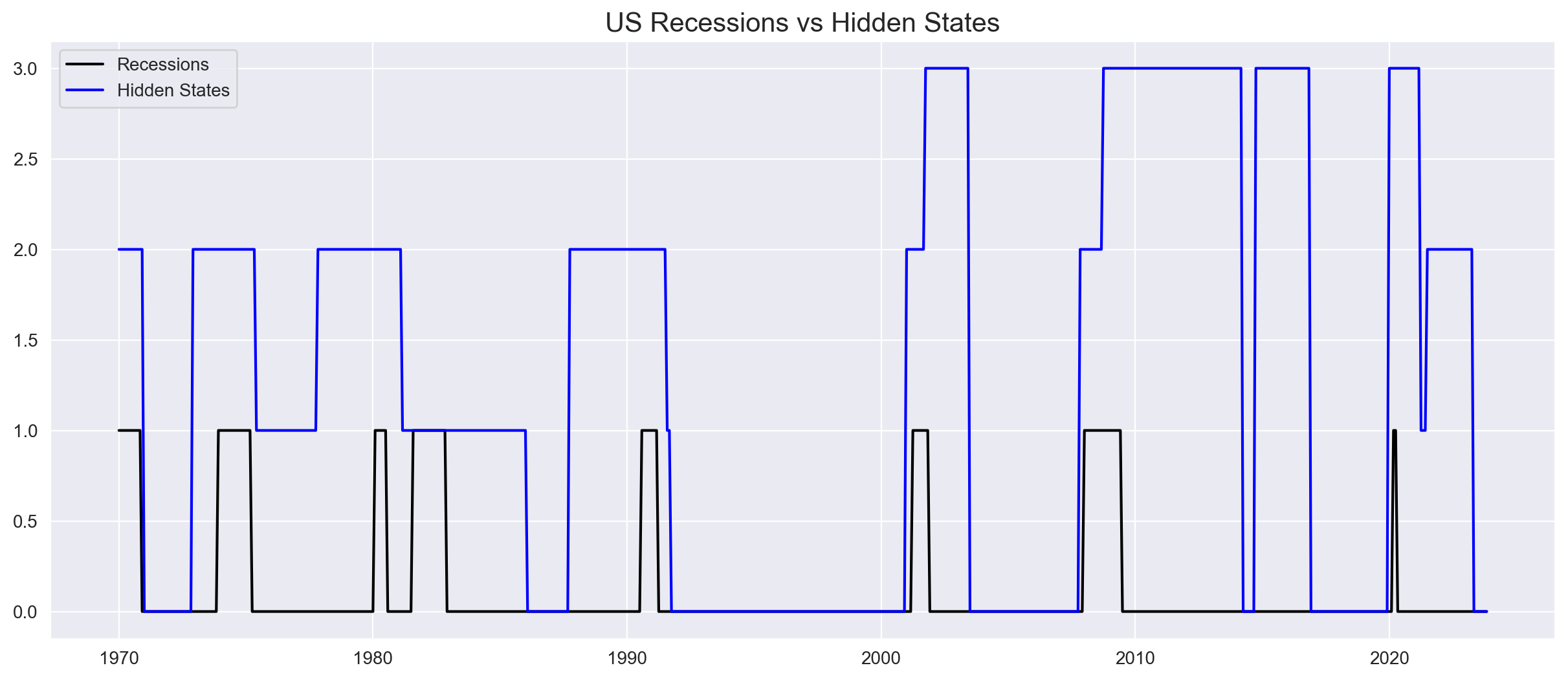}
    \caption{HMM to Recessions}
    \label{fig:hmmr}
\end{figure}

To assess the validity of the HMM, \autoref{fig:hmmr} shows the comparison of the hidden states to the NBER recessions. The model is able to capture important recessions such as the 2008 financial crisis, COVID-19, and the oil crisis in the 1970s, and is also able to pick up on smaller economic crises such as the 2015 decline in industrial production and state recessions. To simplify the intuition of the hidden states, we describe them as follows, state 0 represents economic stability, while states 1, 2, and 3 represent an economic crisis.

\begin{figure}[!ht]
    \centering
    \includegraphics[width=0.95\textwidth]{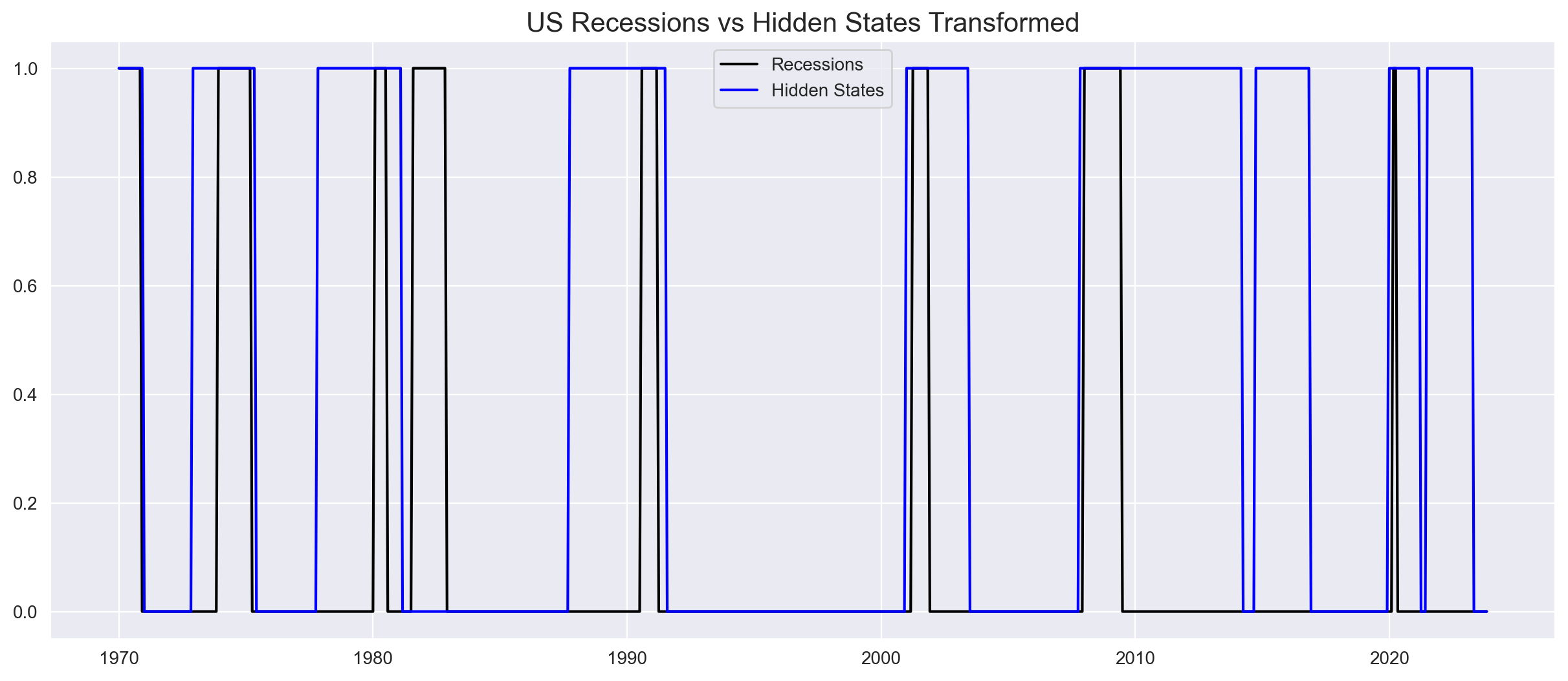}
    \caption{HMM to Recessions Transformed}
    \label{fig:hmmrtf}
\end{figure}

\noindent  Since the NBER recessions are either 0 or 1 a transformation was done to bring the maximum of the hidden states down to one, and can be seen below in \autoref{fig:hmmrtf}.

The hidden states match the recessions relatively well and \autoref{tab:clsrep} is the classification report. The model attains a 62\% accuracy given the severe class imbalance, which suggests good performance, and the precision and recall scores suggest that the model is not just predicting one class and is able to learn how the economy is functioning. Unfortunately, standard transformations such as down or upsampling cannot be done to fix the class imbalance, since they would not be an accurate representation of the US economy. The HMM returns the hidden states and the means for each of those states by feature, and \autoref{fig:mhmm} is available in the appendix and visualizes the hidden states and means at each state for each variable. 

\begin{table}[!ht]
    \centering
    \begin{adjustbox}{width=0.5\textwidth}
    \begin{tabular}{rrrrr}
\toprule
 & precision & recall & f1-score & support \\
\midrule
0.0 & 0.95 & 0.59 & 0.73 & 562 \\
1.0 & 0.23 & 0.81 & 0.36 & 85 \\
\\
accuracy & & & 0.62 & 647 \\
macro avg & 0.59 & 0.70 & 0.55 & 647 \\
weighted avg & 0.86 & 0.62 & 0.68 & 647 \\
\bottomrule
\end{tabular}

    \end{adjustbox}
    \caption{Classification Report}
    \label{tab:clsrep}
\end{table}

\subsection{HMM with Oil Price Data} \label{section:oilprice}
\begin{figure}[!ht]
    \centering
    \includegraphics[width=0.95\textwidth]{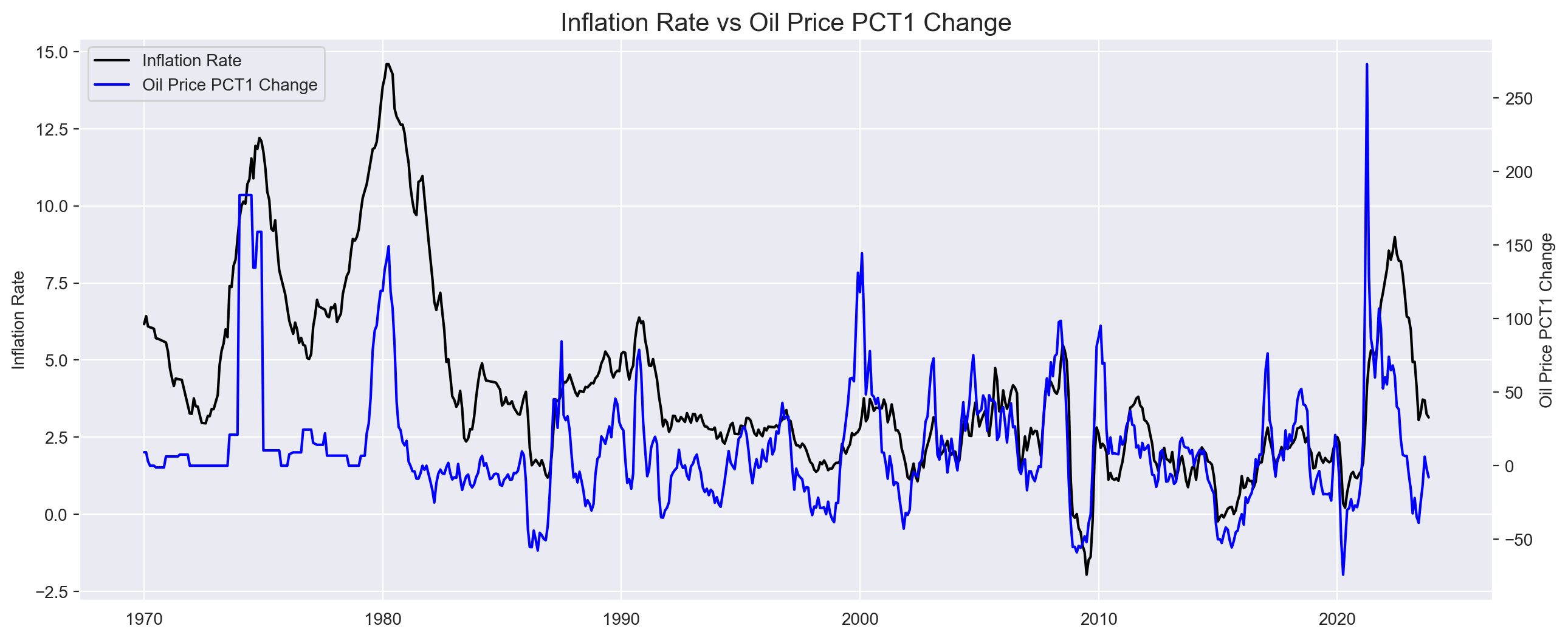}
    \caption{Inflation Rate vs. Oil Price}
    \label{fig:inflatoil}
\end{figure}

To ensure that all factors of the economy were covered, WTI crude oil price was considered as a feature. \autoref{fig:inflatoil} shows a comparison of the CPI Inflation Rate and percent change in oil price, where the right y-axis is the oil price, and the left y-axis is the CPI inflation rate. 

\begin{figure}[!ht]
    \centering
    \includegraphics[width=0.95\textwidth]{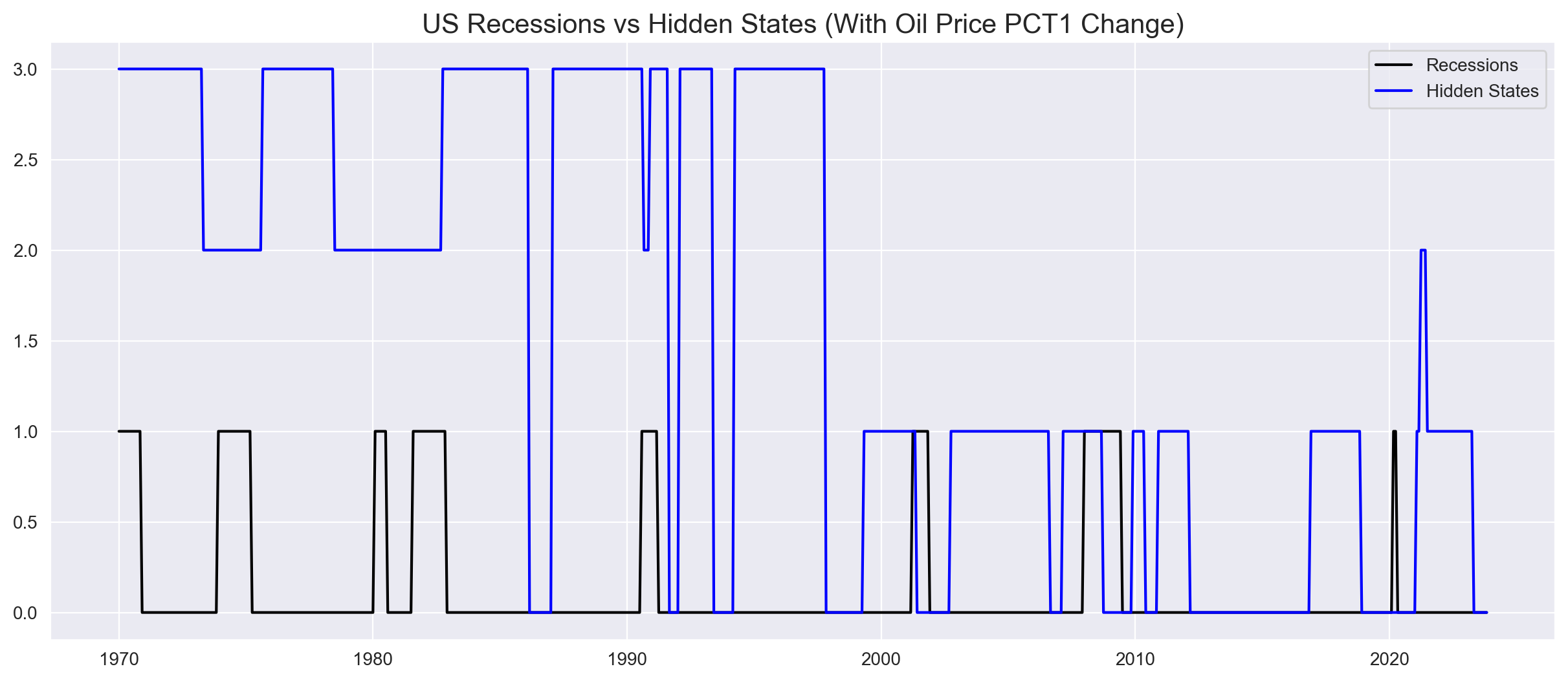}
    \caption{HMM (With Oil) to Recessions}
    \label{fig:hmmoil}
\end{figure}

\noindent Oil price is extremely volatile as seen above, and though it follows the same trend as inflation, due to its range it is given more weight and importance in the selected models. \autoref{fig:hmmoil}, shows the performance of the HMM with the oil price data as a feature. A diagonal covariance matrix was used for this as well to ensure consistency, and visually the HMM does not capture the recessions as effectively.

\noindent The decision to omit oil price data was made to streamline the modeling process, prioritizing factors with more stable and consistent effects on inflation. Focusing on more stable indicators enhances the robustness of the forecasting models, making them better equipped to handle various economic conditions and changes in market dynamics over time. 

Since oil price was excluded, the hidden states and the means derived by the HMM, from \autoref{fig:hmmr}, without oil price, were added as features to the original dataset creating three new datasets: original data with hidden states as a new feature, original data with means as new features, and original data with hidden states and means as new features. Integrating these derived features into our modeling approach, can enhance the predictive capabilities of other models and gain a deeper understanding of the economic factors influencing inflation rates.

\newpage
\section{Neural Networks}
Neural networks are brain-inspired supervised learning models made with interconnected nodes and layers to learn patterns and make predictions from data. Unlike linear models, they are able to capture non-linear relationships between the target and features; however, this leads to their biggest drawback, interpretability. Neural networks are black-box models, which means that the user only sees what goes in and what comes out, which makes them difficult to interpret compared to traditional models. Neural networks have seen a rise in usage in Economics, but due to their lack of interpretability, they get overshadowed by simpler models. There are different types of neural networks, and this paper focuses on Recurrent Neural Networks (RNN), specifically Long Short-Term Memory (LSTM) networks.

\subsection{Artificial and Recurrent Neural Networks}
Artificial neural networks (ANN) are the simplest type of neural networks. They consist of multiple layers and neurons and can be used for regression and classification tasks. Each node has an activation function that transforms the incoming input and passes it on to the next layer. Common activation functions are sigmoid, tanh, ReLU, and linear, and depending on the task, different functions are picked. Training an ANN involves adjusting weights using algorithms like backpropagation to minimize the difference between predicted and actual outputs. 

RNNs introduce recurrent connections, allowing information to persist and be shared across different time steps in sequential data, where temporal dependencies are crucial. These models are able to handle variable-length sequences, allowing them to adapt dynamically to different lengths of input data. Their architectures can be extended and optimized with variations such as Long Short-Term Memory (LSTM) and Gated Recurrent Units (GRU), which improve the model's ability to retain and utilize information over longer sequences.

\subsection{Long Short-Term Memory (LSTM) Models}
\begin{figure}[!ht]
    \centering
    \includegraphics[width=0.5\textwidth,height=1.8in]{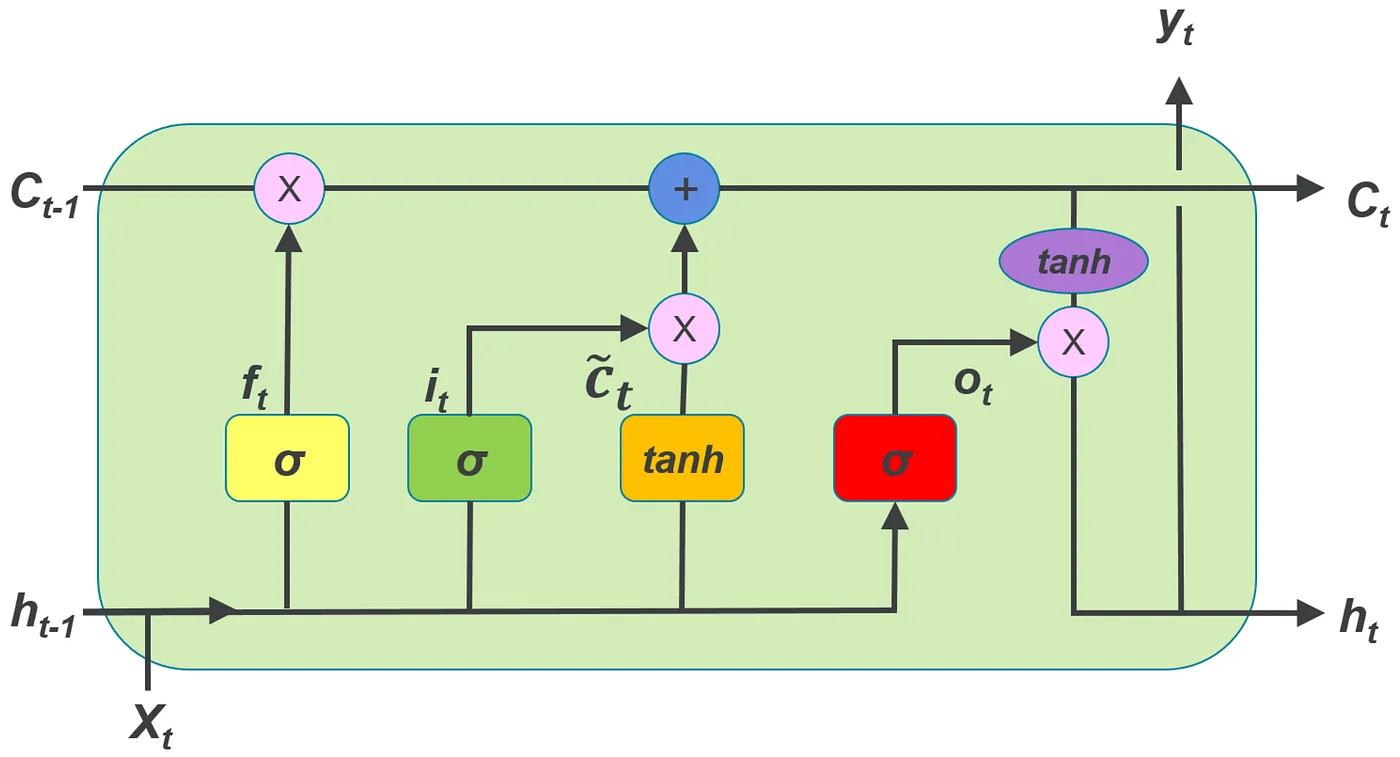}
    \caption{LSTM Neuron at any time step $t$ \citep{lstmneuron}}
    \label{fig:LSTM}
\end{figure}

LSTMs are a type of recurrent neural network and function very similarly to the human brain. They are designed to address the vanishing gradient problem in traditional RNNs, allowing them to better capture time dependencies in sequential data, and they achieve this through the use of gated units called memory cells that can maintain information over time. \autoref{fig:LSTM} represents a single LSTM neuron, and shows how the information is processed.

\noindent Memory cells with input, forget, and output gates regulate the flow of information, allowing the model to selectively remember or forget information. They are trained using backpropagation through time (BPTT), adjusting the weights to minimize the error in the predictions.

\autoref{eq:lstmeq} from \citep{lstm} is the matrix representation for the LSTM cell, and unlike regular neurons in a neural network, LSTM neurons use tanh as the main activation function and sigmoid as the recurrent activation function. 
\begin{align}\label{eq:lstmeq}
\begin{pmatrix}[0.6]i \\ f \\ o \\ g\end{pmatrix} &= \begin{pmatrix}[0.6]\sigma \\ \sigma \\ \sigma \\ \tanh \end{pmatrix} W \begin{pmatrix}[0.6]h_{t-1}\\ x_{t}\end{pmatrix}\\
h_{t} &= o_{t}\odot \tanh (f \odot c_{t-1} + i \odot g) \notag
\end{align}
\vspace{-0.5mm}
For this research, the LSTM architecture was defined as follows: Input layer, 2 LSTM layers (tanh and sigmoid activation), 1 Dense Layer (linear activation), and the output layer.

\newpage
\subsection{In-Sample Forecasting}
With the defined architecture, the model was first tested on a 70:30 split of the data. This particular split was chosen as it allowed for one recession and one expansion to be in the training and testing set. The training set ends at the bottom of the 2008 recession, and the test set starts at the bottom of the 2008 recession to the present. Before training the model, preprocessing of the data was done to ensure numerical stability. Since LSTMs are sensitive to the input data, MinMaxScaler, defined in \autoref{eq:mms}, was used to scale the values in the columns to be between 0 and 1.
\begin{align}\label{eq:mms}
    x' = a + \frac{(x - \min (x))(b - a)}{\max (x) - \min (x)}, a = 0 \text{ and } b=1
\end{align}
The scaler ensured that all features were uniformly scaled, preventing any single variable from dominating the model's learning process due to differences in scale. To capture the time-dependent information, 24 lags were used on all the variables and transformed into a 3D matrix (tensor) for the LSTM.

\begin{figure}[!ht]
    \centering
    \includegraphics[width=0.95\textwidth]{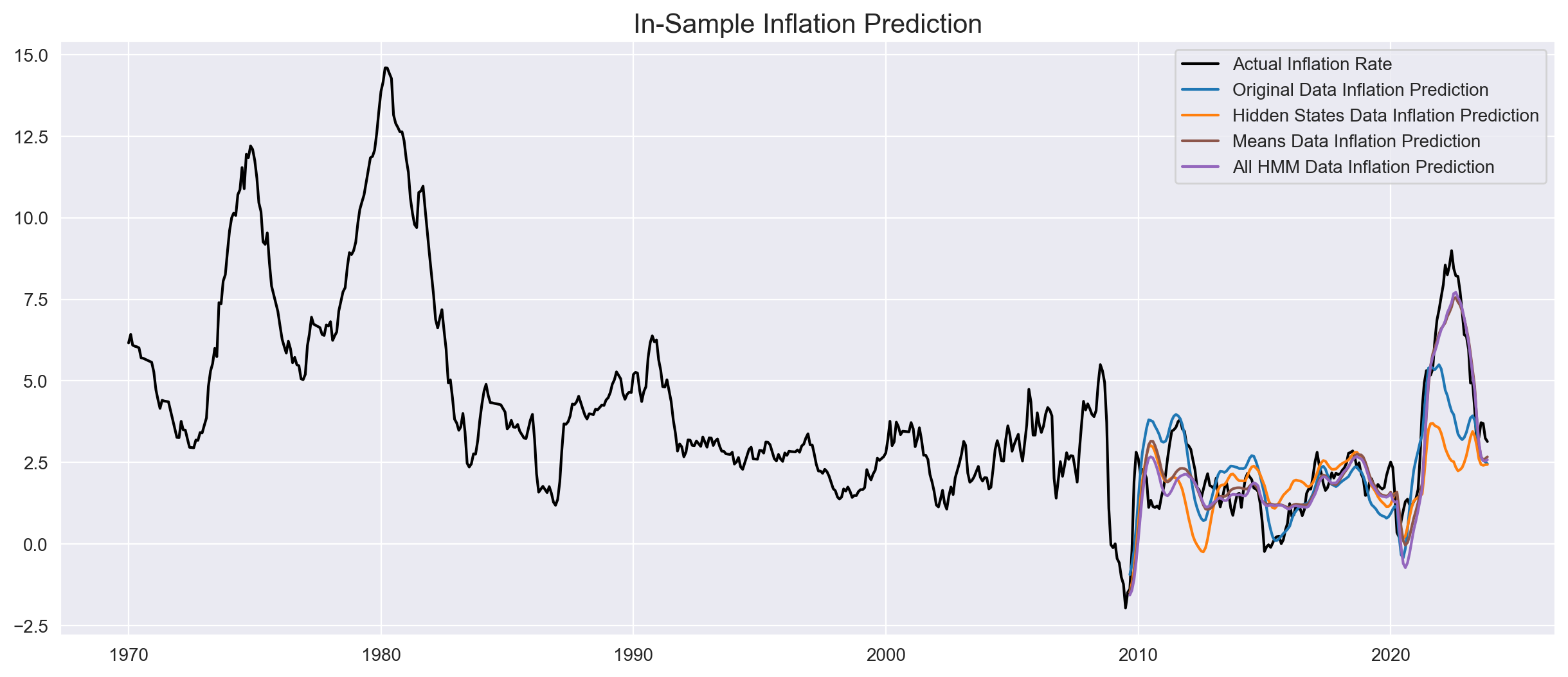}
    \caption{In-Sample Inflation Prediction}
    \label{fig:inslstm}
\end{figure}

After preprocessing, the model was tested on these 4 datasets, original data, original data with hidden states as a feature, original data with means as features, and original data with hidden states and means as features. Their performance is shown below in \autoref{tab:metrep}
\begin{table}[!ht]
    \centering
    \begin{adjustbox}{width=0.6\textwidth}
    \begin{tabular}{lrrrr}
\toprule
 & Original Data & Hidden State Data & Means Data & All HMM Data \\
\midrule
MSE          & 1.992 & 3.548 & 0.818 & 0.861 \\
MAE          & 0.964 & 1.290 & 0.694 & 0.705 \\
$\text{R}^2$ & 0.519 & 0.144 & 0.803 & 0.792 \\
\bottomrule
\end{tabular}
    \end{adjustbox}
    \caption{Metrics Report}
    \label{tab:metrep}
\end{table}
and we can see a significant improvement in all 3 metrics by adding the means as features to the dataset. This improvement indicates that the LSTM model effectively leveraged the additional information provided by the means, allowing it to better capture the underlying time dependencies present in the data. Visually, we can see the improvement in \autoref{fig:inslstm}, where the data with the means is able to predict the rise and fall in 2022. 

\noindent Unfortunately, due to the minimal rows, only a train-test split was done, and the training loss curves for the different predictions can be found in \autoref{fig:losscurve}; we can see a smooth decrease with no sharp turns, suggesting that the model is not overfitting. Compared to traditional methods, the LSTM forecast is able to capture the temporal dependencies and predict CPI inflation rates.

\subsection{Feature Importance using Integrated Gradients}
With the increase in popularity of Explainable AI, Integrated Gradients (IG) have become popular a technique for feature importance for complex models such as neural networks. They are commonly used in image classification tasks, but since they work for any differentiable model, it has been adapted to interpret time series data. IG works by computing the integral of the gradient of a baseline input and the actual input, a Riemann approximation is used and is given below in \autoref{eq:igeq} from \citep{igequation}:
\begin{align}\label{eq:igeq}
IG_{i}(x) \approx (x_{i}-x_{i}') \times \sum\limits_{k=1}^{m} \frac{\partial F \left(x' + \frac{k}{m} \times (x-x') \right)}{\partial x_{i}} \times \frac{1}{m}
\end{align}

\begin{figure}[!ht]
    \centering
    \includegraphics[width=0.9\textwidth]{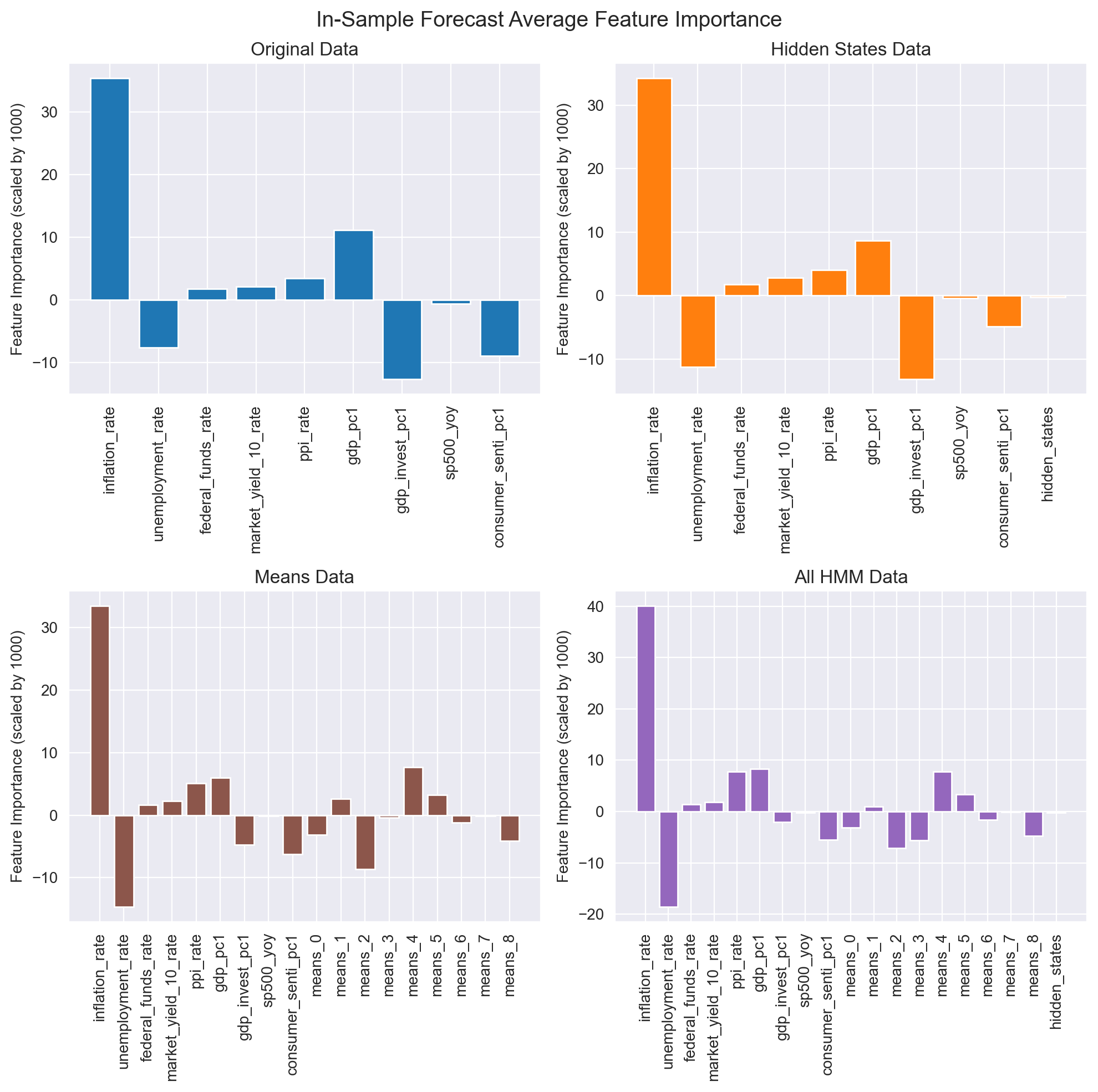}
    \caption{In-Sample Forecast Average Feature Importance}
    \label{fig:isfi}
\end{figure}

Since the IG computation can get very computationally expensive, $m$ was set to 50, so the values are just an approximation. However, they still provide insights into how the LSTM is able to learn and determine the relationships between the economic variables. \autoref{fig:isfi} is the integrated gradients, averaged across the test set and lags, for the different features. The LSTM is able to correctly determine the inverse relationship between the inflation rate and unemployment rate, and the relationship with the other features follows the economic theory. By adding the means as features, it increases the importance of the lags of inflation and the other variables which could be leading to the prediction improvements. 

\newpage
\subsection{Time Series Cross Validation}
To assess the LSTM model across different time periods, forward chaining cross-validation was conducted. Since there are only around 650 rows of data, and neural networks require a lot of data, 5 folds were chosen. \autoref{tab:metrepcv} provides a summary of the metrics across the folds
\begin{table}[!ht]
    \centering
    \begin{adjustbox}{width=0.6\textwidth}
    \begin{tabular}{llrrrrr}
\toprule
 & & Fold 1 & Fold 2 & Fold 3 & Fold 4 & Fold 5\\
\midrule

\multirow{3}{*}{Original Data} & MSE & 11.580 & 0.486 & 0.490 & 1.080 & 1.631\\
& MAE & 2.930 & 0.597 & 0.549 & 0.740 & 1.008\\
& $\text{R}^2$ & -0.911 & 0.590 & 0.330 & 0.509 & 0.711\\
\midrule

\multirow{3}{*}{Hidden State Data} & MSE & 11.895 & 1.046 & 0.623 & 0.862 & 0.799\\
& MAE & 3.080 & 0.884 & 0.629 & 0.672 & 0.677\\
& $\text{R}^2$ & -0.963 & 0.118 & 0.149 & 0.608 & 0.858\\
\midrule

\multirow{3}{*}{Means Data} & MSE & 10.044 & 1.125 & 1.152 & 1.223 & 0.903\\
& MAE & 2.728 & 0.891 & 0.786 & 0.794 & 0.716\\
& $\text{R}^2$ & -0.657 & 0.052 & -0.574 & 0.444 & 0.840\\
\midrule

\multirow{3}{*}{All HMM Data} & MSE & 8.694 & 1.522 & 1.390 & 1.360 & 0.694\\
& MAE & 2.566 & 1.018 & 0.823 & 0.838 & 0.632\\
& $\text{R}^2$ & -0.435 & -0.283 & -0.901 & 0.382 & 0.877\\

\bottomrule
\end{tabular}
    \end{adjustbox}
    \caption{Metrics Report (5 Fold CV)}
    \label{tab:metrepcv}
\end{table}

\begin{figure}[!ht]
    \centering
    \includegraphics[width=0.95\textwidth]{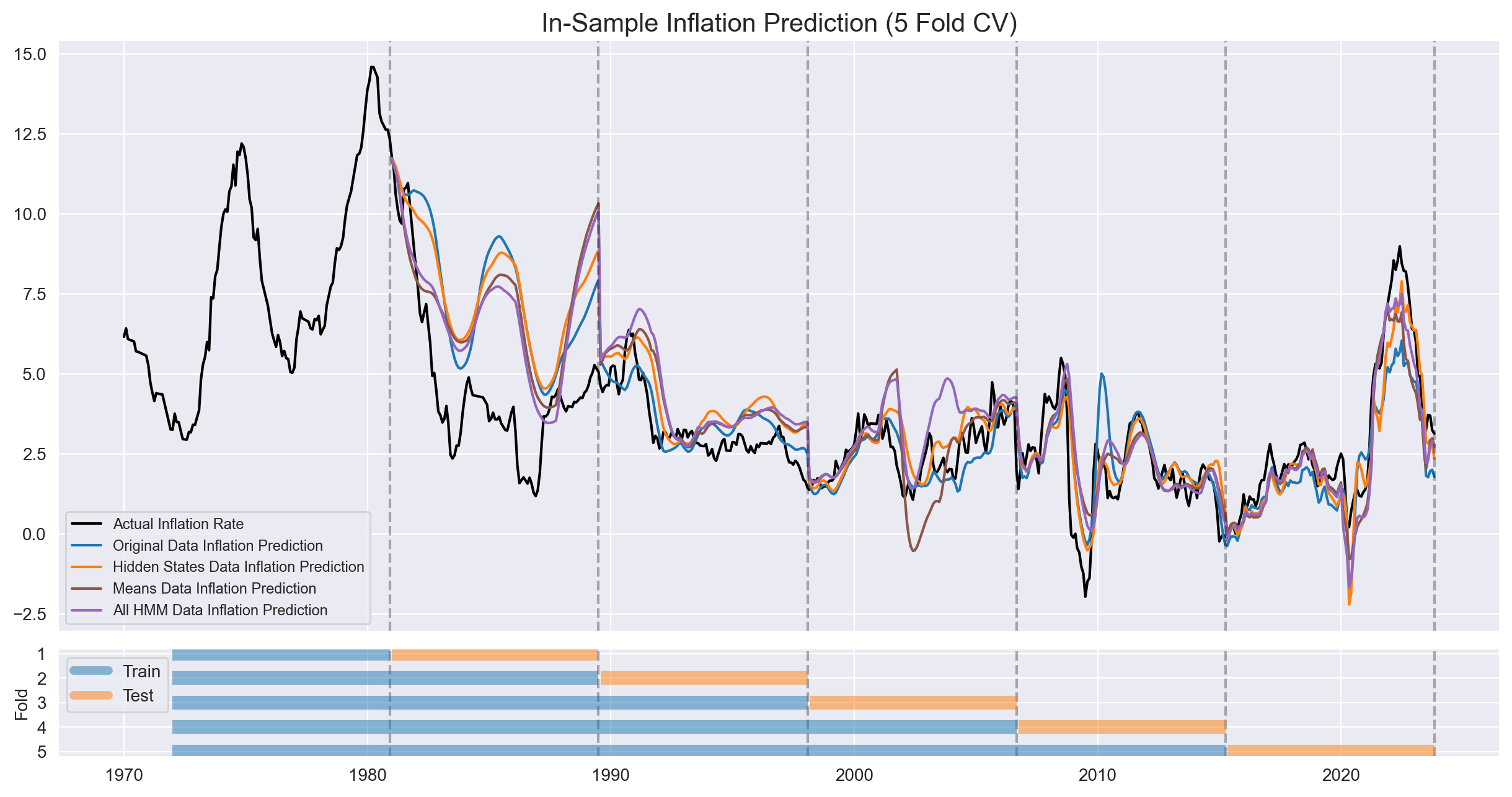}
    \caption{In-Sample Inflation Prediction (5 Fold CV)}
    \label{fig:inslstmcv}
\end{figure}

\noindent We can see an improvement in the metrics with each fold which makes sense as the model has more data to learn from. \autoref{fig:inslstmcv} is shows the predicted inflation rates 

With 5 folds, COVID and the 2008 financial crisis were given their own fold, allowing for a comprehensive assessment of the model's generalizability during economic crises. We can see in the plot that the model is able to predict the recession, and over time the model performs better. The plot of the loss curves are in \autoref{fig:losscurvecv}, and across folds demonstrated consistent performance, although fold 4 exhibited slightly higher variability, possibly indicating that the 2008 recession was difficult for the LSTM to capture. Unfortunately, Integrated Gradients were not feasible for cross-validation (CV) due to computational constraints. Additionally, while hyperparameter tuning and regularization were conducted, they were not extensively explored for the same reason. 

LSTMs prove to be powerful tools for time series forecasting, leveraging recurrent connections to capture temporal dependencies effectively. The observed improvement emphasizes the synergy between HMM-derived features and LSTM modeling, highlighting the value of integrating diverse modeling approaches. The fusion enhances predictive accuracy and also offers valuable insights into economic dynamics, allowing us to extend this approach to forecast future inflation rates.

\newpage
\section{Out-of-Sample Forecasting}
\subsection{1-year Forecast}
\begin{figure}[!ht]
    \centering
    \includegraphics[width=0.95\textwidth]{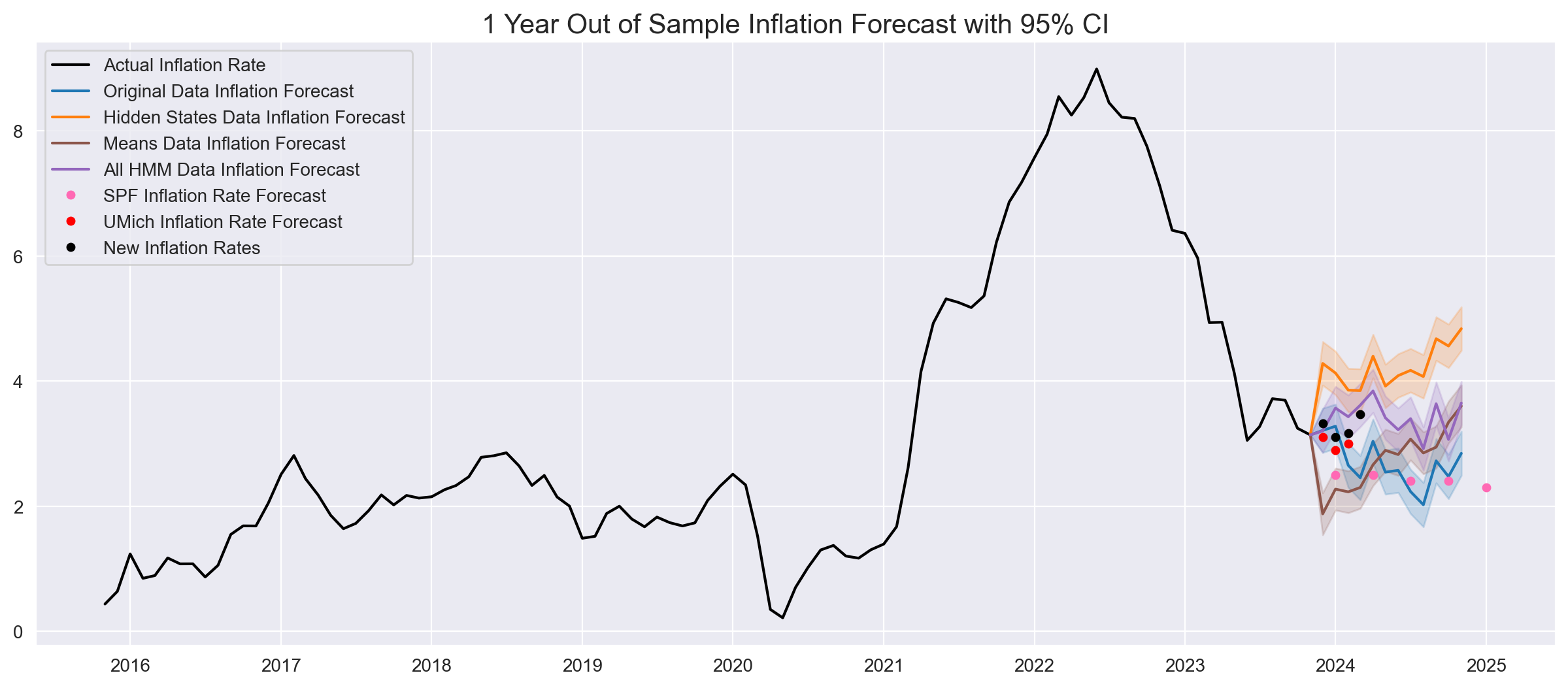}
    \caption{1 Year Out-of-Sample Forecast}
    \label{fig:osf1}
\end{figure}

Following the in-sample forecast, future forecasts were conducted to predict the inflation rate. The output layer of LSTM architecture was slightly modified to account for the forecast horizon. For the 1 year forecast, the output layer consisted of 12 neurons, one for each month. LSTMs are able to handle multiple features and account for them in the forecast, which makes them valuable in economics. To ensure that enough past data was captured, without overfitting, a total of 48 lags were used. The forecast with their confidence intervals is shown in \autoref{fig:osf1}

\begin{figure}[!ht]
    \centering
    \includegraphics[width=0.9\textwidth]{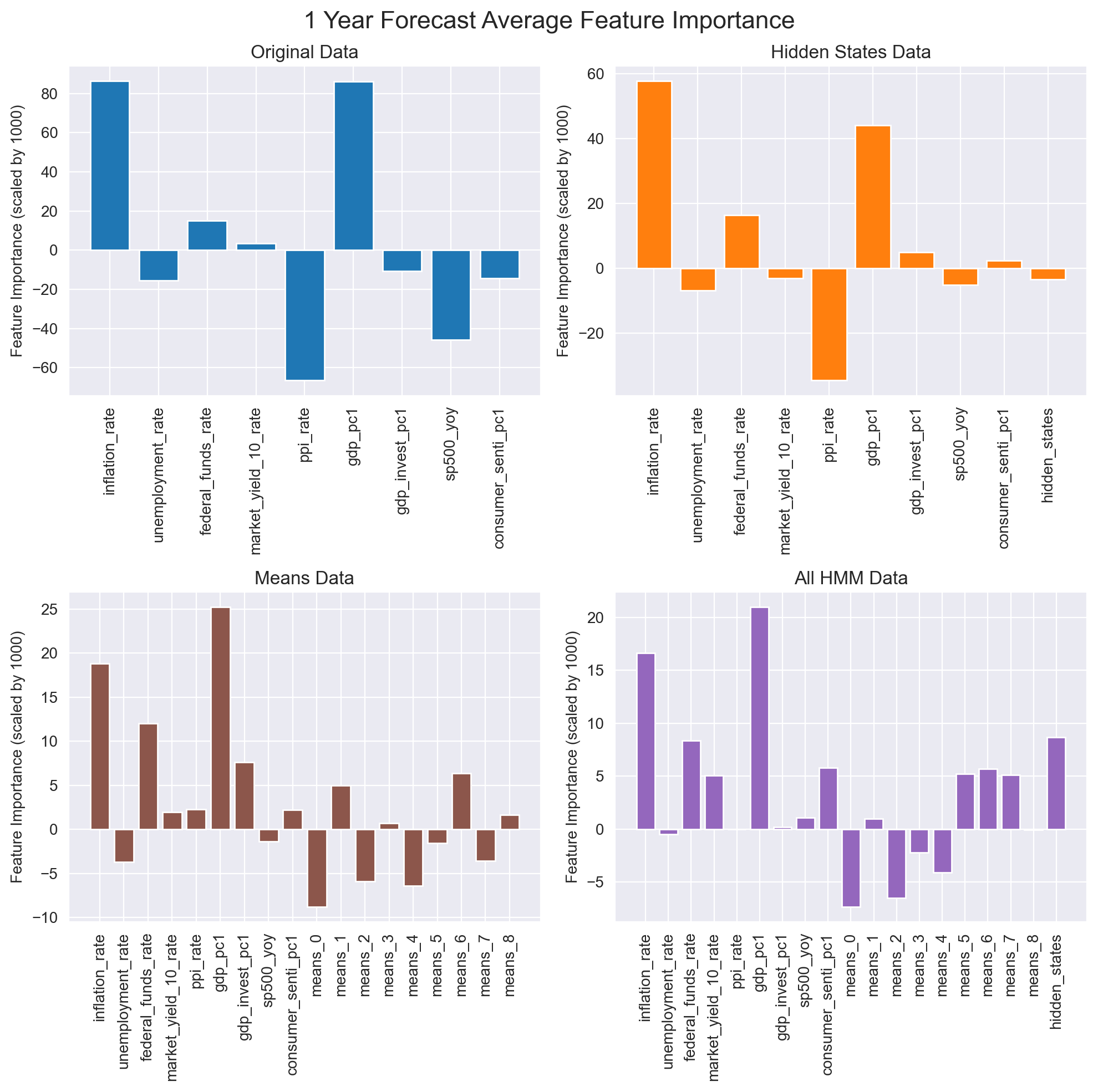}
    \caption{1 Year Forecast Average Feature Importance}
    \label{fig:osfi1}
\end{figure}

Based on the in-sample performance, the model with all the data was expected to perform the best, and it is seen when comparing it to new inflation rate values, as they fall in the confidence interval. Since that data was collected up until November 2023, the following inflation rates were used as a benchmark. The model was able to predict the March 2024 inflation rate while official forecasts had predicted a decline. Incorporating the HMM-derived features provided valuable insights into economic conditions, enhancing the predictive capability of the model and contributing to its accurate forecasting performance. To visualize the feature importance, the IG is provided below in \autoref{fig:osfi1}

Following economic theory, GDP growth has the most importance in forecasting future inflation rate, followed by the lags of inflation. The means tend to have an inverse relationship with the forecast, and this could be leading to its improved performance. Comparing it to using the original data, the importance of the features is reduced significantly, but the means are adding valuable information that is otherwise not captured. 

\newpage
\subsection{2.5-year Forecast}
\begin{figure}[!ht]
    \centering
    \includegraphics[width=0.95\textwidth]{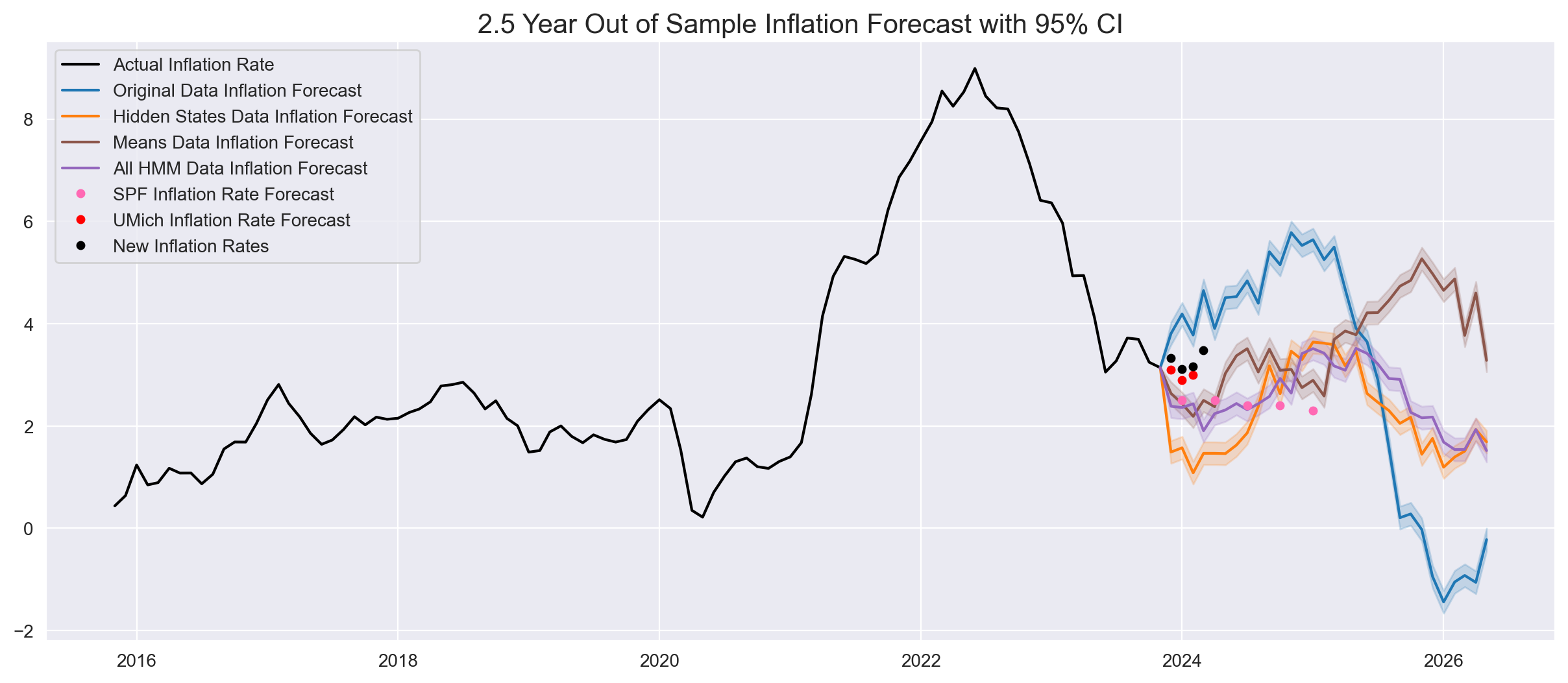}
    \caption{2.5 Year Out-of-Sample Forecast}
    \label{fig:osf25}
\end{figure}

To test the effectiveness of the model, a longer-term forecast was conducted. Similar to the 1-year forecast, the output layer of the LSTM was adjusted to 30 neurons, and 120 lags were used since enough information needed to be captured. Unfortunately, the forecast was not as good as the 1-year forecast, seen in \autoref{fig:osf25}, but the addition of the HMM features led to a less volatile forecast when compared to the original data. 

\begin{figure}[!ht]
    \centering
    \includegraphics[width=0.9\textwidth]{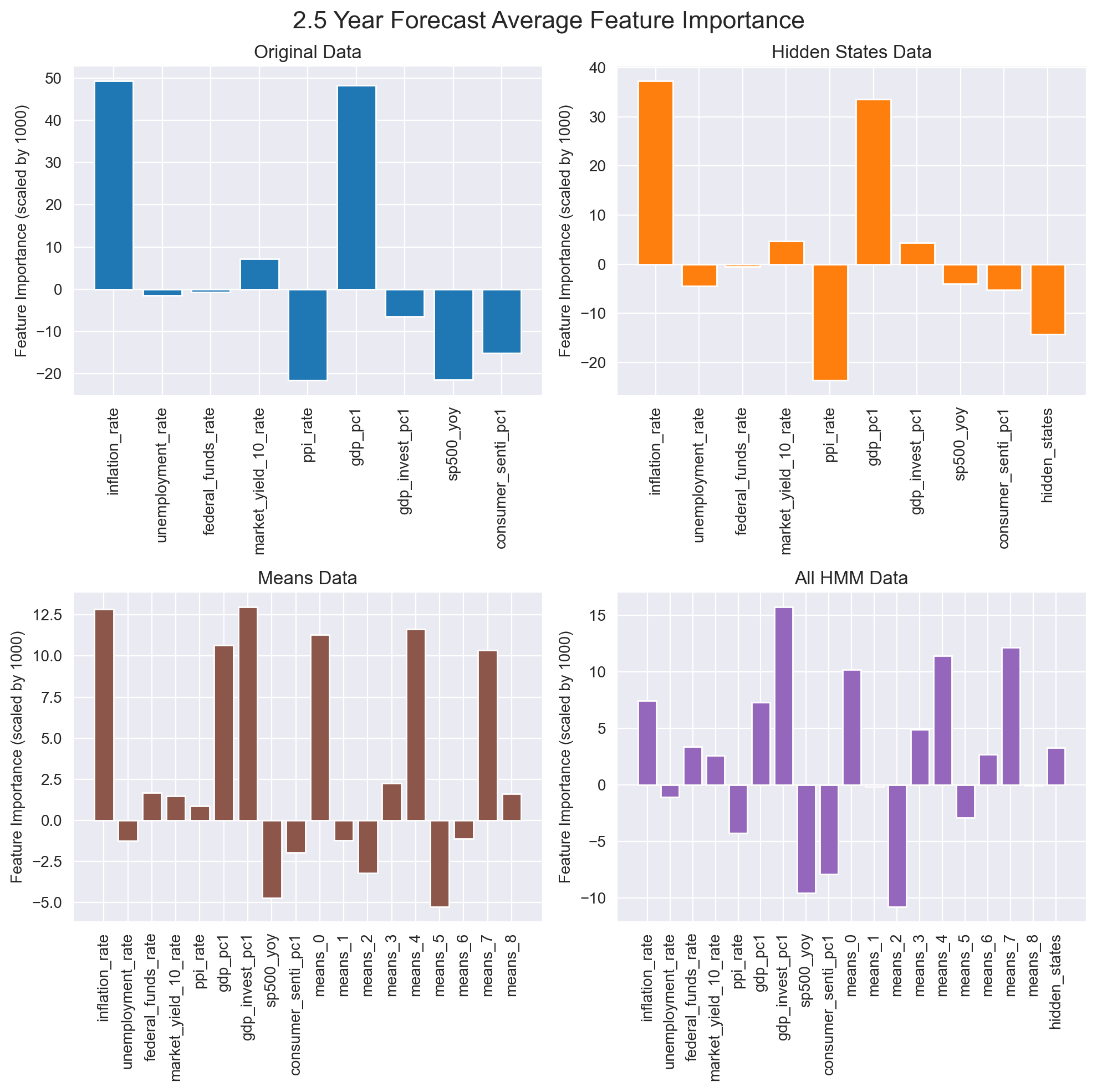}
    \caption{2.5 Year Forecast Average Feature Importance}
    \label{fig:osfi25}
\end{figure}

The decrease in volatility could be explained by the fact that averaging is a common technique used to reduce variance. By incorporating the HMM means, the inflation rate forecast adjusts accordingly and does not predict a severe recession as indicated by the original data. Though none of the curves match the new inflation rate values, the overall average across the 30 months was around 3\%, which has been the average for the last 10-20 years. To further enhance the forecast accuracy, various strategies were explored, including experimenting with different lag lengths and conducting thorough hyperparameter tuning. Despite these efforts, the model's performance for the 2.5-year forecast fell short compared to the 1-year forecast. One of the interesting observations from the IG, in \autoref{fig:osfi25}, was that GDP investment was given significantly more importance compared to the 1-year forecast.

GDP investment growth is a leading indicator, so it tends to react before the inflation rate changes. The model could potentially suggest that for longer-term forecasts, more weight should be given to GDP investment compared to the other variables. Additionally the year-over-year returns of SP500 were also significant, though inversely related, while the means of SP500 year-over-year returns have a positive relationship. While the 2.5-year forecast did not achieve the same level of performance as the 1-year forecast, insights gained from feature importance analysis shed light on potential adjustments to improve future forecasting.

\newpage
\section{Future Work}
The results from this paper provide promising results; however, there are several avenues for future research. Particularly, GDP and GDP Investment were interpolated, so utilizing the exact values would be more appropriate and could potentially lead to better results. As mentioned earlier, the Integrated Gradients are just an approximation, so with a more powerful machine, more steps could lead to providing more detail into the feature importance. In addition, methods such as Shapley and LIME could be modified to extract insights into recurrent neural networks since they have been proven to perform better than integrated gradients. Similarly, extensive hyperparameter tuning could lead to improved performance in both the HMM and LSTM. 

With recent advancements in time series forecasting, different models could be tested such as Temporal Fusion Transformers (TFT), models that use a transformer architecture to obtain an interpretable multivariate forecast. Despite their current computational demands and dataset-specific applicability, advancements in hardware and algorithmic efficiency may soon render TFTs viable for forecasting a wide array of economic variables. Investigating the adaptability and scalability of TFTs within the context of economic forecasting will lead to new insights and predictive capabilities.

Furthermore, exploring ensemble methods that combine the strengths of different forecasting approaches, such as HMMs and LSTMs, could offer further improvements in predictive accuracy and robustness. A Markov switching approach, similar to Markov Switching Autoregressions, could be applied to Neural Networks to create Markov Switching Neural Networks. Instead of just augmenting the features to the dataset, this approach would involve integrating a Markov layer within the neural network architecture to autonomously capture regime switches. By leveraging the complementary strengths of multiple models, the forecasts can navigate the complexities of economic dynamics.

\newpage
\section{Conclusion}
This paper presented an interpretable inflation forecasting with a Hidden Markov Model (HMM) and a Long Short-Term Memory (LSTM) fusion model with integrated gradients. The study aimed to explore the effectiveness of incorporating HMM features into LSTM models for inflation prediction, leveraging the dynamic nature of economic data. The fusion model provided valuable insights into the predictive capabilities to capture complex relationships within various economic variables.

The integration of HMM features, such as hidden states and means, significantly enhanced the predictive performance of LSTM models for forecasting. By utilizing the latent state information provided by HMMs, the LSTM models were able to capture subtle shifts in the economy, resulting in more accurate and robust predictions. While the models demonstrated strong performance in short-term predictions, their effectiveness over longer forecasting horizons was somewhat limited. Further modeling, refinement, and experimentation with different economic features could be done to create stable long-term forecasts.

The additional insights gained from the integrated gradients suggest that the model is able to capture the complex dynamics found in economic theory. The augmentation of the HMM-derived data further enhanced the model to determine those relationships and produce reliable forecasts. Compared to the official forecasts and expectations, the fusion model can accurately capture new trends in the 1-year forecast. 

With probabilistic modeling, the model was able to capture the inherent uncertainty in economic data, enabling a more robust analysis of inflation trends. Meanwhile, the integration of deep learning with integrated gradients allowed for the extraction of complex patterns and relationships, facilitating deeper insights into the drivers of inflation dynamics. The fusion of the two models, with integrated gradients, offers a new technique for interpretable and accurate inflation forecasting.

\newpage
\singlespacing
\nocite{scikit-learn, tensorflow, spf, githubcode}
\bibliography{references_honors}
\addcontentsline{toc}{section}{References}

\newpage
\section*{Appendix A: Tables}
\addcontentsline{toc}{section}{Appendix A: Tables}

\renewcommand{\thetable}{A\arabic{table}}
\setcounter{table}{0}

\begin{table}[!ht]
    \centering
    \begin{adjustbox}{width=1\textwidth}
    \begin{tabular}{lrrrrrrrrrr}
\toprule
 Variable & ADF-Statistic & P-Value & Used Lag & Nobs & 1\% & 5\% & 10\% & Stationary\\
\midrule
inflation\_rate & -2.9505 & 0.0398 & 15 & 631 & -3.4408 & -2.8661 & -2.5692 & Yes\\
unemployment\_rate & -3.3240 & 0.0138 & 2 & 644 & -3.4405 & -2.8660 & -2.5692 & Yes\\
federal\_funds\_rate & -2.5860 & 0.0959 & 17 & 629 & -3.4408 & -2.8661 & -2.5692 & No\\
market\_yield\_10\_rate & -1.4700 & 0.5483 & 12 & 634 & -3.4407 & -2.8661 & -2.5692 & No\\
ppi\_rate & -5.1161 & 0.0000 & 15 & 631 & -3.4408 & -2.8661 & -2.5692 & Yes\\
gdp\_pc1 & -4.5632 & 0.0002 & 12 & 634 & -3.4407 & -2.8661 & -2.5692 & Yes\\
gdp\_invest\_pc1 & -4.7711 & 0.0001 & 12 & 634 & -3.4407 & -2.8661 & -2.5692 & Yes \\
sp500\_yoy & -4.9281 & 0.0000 & 13 & 633 & -3.4407 & -2.8661 & -2.5692 & Yes\\
consumer\_senti\_pc1 & -5.4568 & 0.0000 & 16 & 630 & -3.4408 & -2.8661 & -2.5692 & Yes\\
oil\_price\_pc1 & -5.3638 & 0.0000 & 12 & 634 & -3.4407 & -2.8661 & -2.5692 & Yes\\
\bottomrule
\end{tabular}
    \end{adjustbox}
    \caption{ADF Test Results}
    \label{tab:adftest}
\end{table}

\begin{table}[!ht]
    \centering
    \begin{adjustbox}{width=1\textwidth}
    \begin{tabular}{cc}
    Original Data Model & Hidden States Data Model \\
     \begin{tabular}{lll}
        \toprule
        Layer (type) & Input Shape & Output Shape\\
        \midrule
        lstm\_input & (None, 24, 9) & (None, 24, 9)\\
        lstm (LSTM) & (None, 24, 9) & (None, 24, 50)\\
        lstm (LSTM) & (None, 24, 50) & (None, 50)\\
        dense (Dense) & (None, 50) & (None, 25) \\
        dense (Dense) & (None, 25) & (None, 1)\\
        \bottomrule
        \end{tabular} &
     \begin{tabular}{lll}
        \toprule
        Layer (type) & Input Shape & Output Shape\\
        \midrule
        lstm\_input & (None, 24, 10) & (None, 24, 10)\\
        lstm (LSTM) & (None, 24, 10) & (None, 24, 50)\\
        lstm (LSTM) & (None, 24, 50) & (None, 50)\\
        dense (Dense) & (None, 50) & (None, 25)\\
        dense (Dense) & (None, 25) & (None, 1)\\
        \bottomrule
        \end{tabular}\\\\\\\\
    Means Data Model & All HMM Data Model \\
     \begin{tabular}{lll}
        \toprule
        Layer (type) & Input Shape & Output Shape\\
        \midrule
        lstm\_input & (None, 24, 18) & (None, 24, 18)\\
        lstm (LSTM) & (None, 24, 18) & (None, 24, 50)\\
        lstm (LSTM) & (None, 24, 50) & (None, 50)\\
        dense (Dense) & (None, 50) & (None, 25)\\
        dense (Dense) & (None, 25) & (None, 1)\\
        \bottomrule
        \end{tabular} &
     \begin{tabular}{lll}
        \toprule
        Layer (type) & Input Shape & Output Shape\\
        \midrule
        lstm\_input & (None, 24, 19) & (None, 24, 19)\\
        lstm (LSTM) & (None, 24, 19) & (None, 24, 50)\\
        lstm (LSTM) & (None, 24, 50) & (None, 50)\\
        dense (Dense) & (None, 50) & (None, 25)\\
        dense (Dense) & (None, 25) & (None, 1)\\
        \bottomrule
        \end{tabular}\\\\\\\\
    1 Year Forecast Model & 2.5 Year Forecast Model\\
     \begin{tabular}{lll}
        \toprule
        Layer (type) & Input Shape & Output Shape\\
        \midrule
        lstm\_input & (None, 48, 19) & (None, 48, 19)\\
        lstm (LSTM) & (None, 48, 19) & (None, 48, 50)\\
        lstm (LSTM) & (None, 48, 50) & (None, 50)\\
        dense (Dense) & (None, 50) & (None, 25)\\
        dense (Dense) & (None, 25) & (None, 12)\\
        \bottomrule
        \end{tabular} &
     \begin{tabular}{lll}
        \toprule
        Layer (type) & Input Shape & Output Shape\\
        \midrule
        lstm\_input & (None, 120, 19) & (None, 120, 19)\\
        lstm (LSTM) & (None, 120, 19) & (None, 120, 50)\\
        lstm (LSTM) & (None, 120, 50) & (None, 50)\\
        dense (Dense) & (None, 50) & (None, 25)\\
        dense (Dense) & (None, 25) & (None, 30)\\
        \bottomrule
        \end{tabular}\\
\end{tabular}

    \end{adjustbox}
    \caption{LSTM Model Summaries}
    \label{tab:modsums}
\end{table}

\newpage
\section*{Appendix B: Figures}
\addcontentsline{toc}{section}{Appendix B: Figures}

\renewcommand{\thefigure}{B\arabic{figure}}
\setcounter{figure}{0}

\begin{figure}[!ht]
    \centering
    \includegraphics[width=0.95\textwidth]{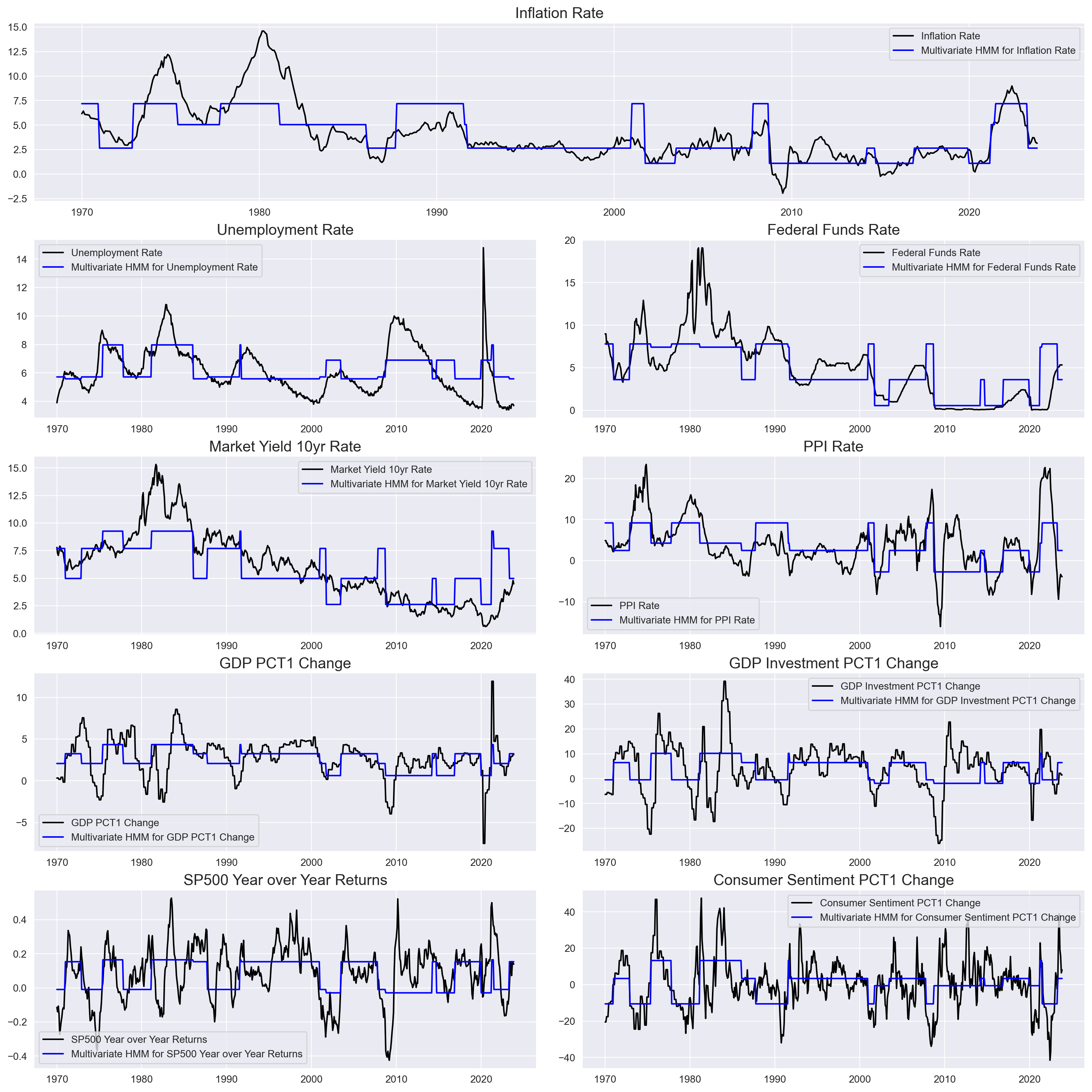}
    \caption{Averages Across Hidden States of the Features}
    \label{fig:mhmm}
\end{figure}

\begin{figure}[!ht]
    \centering
    \includegraphics[width=0.72\textwidth]{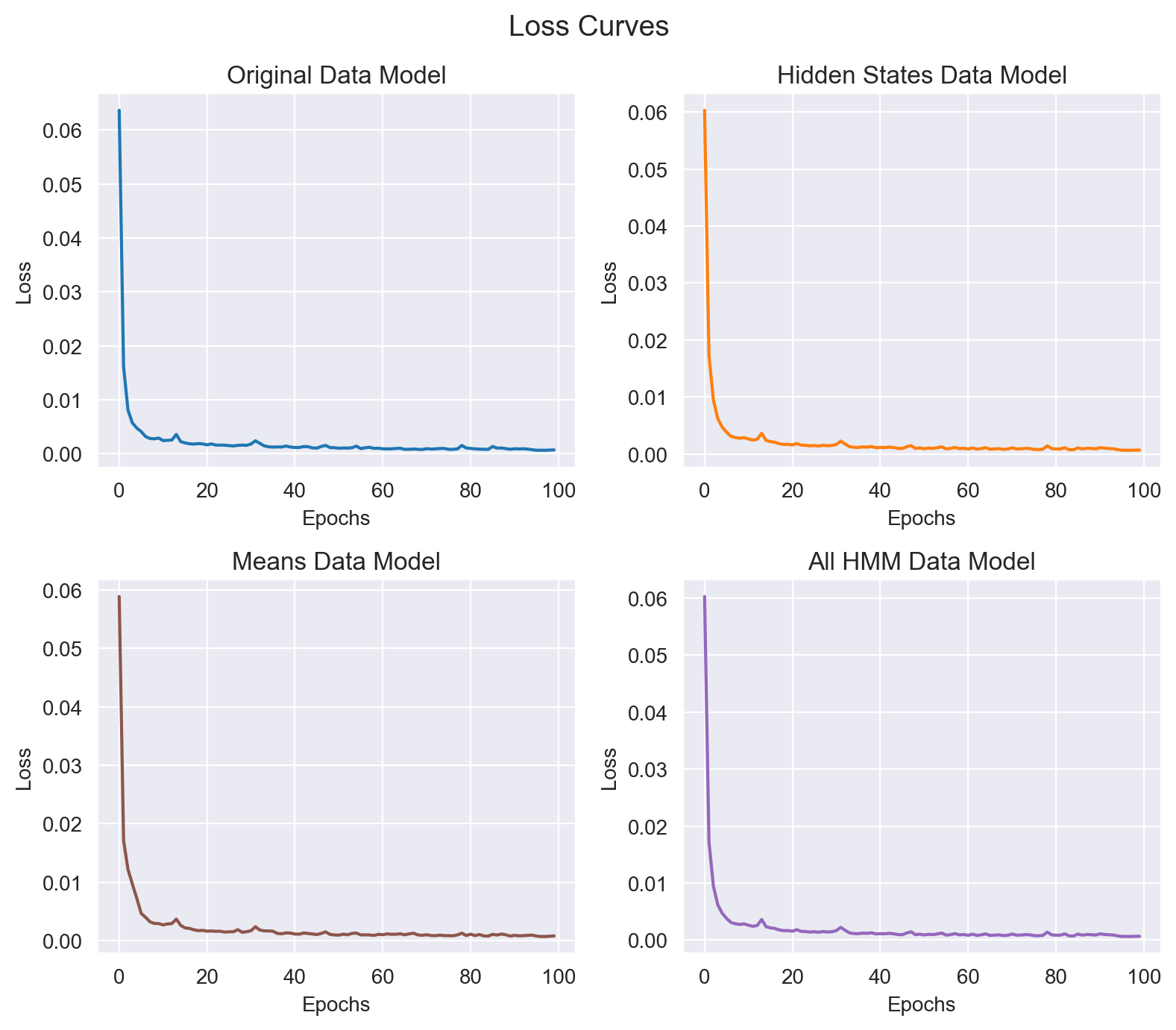}
    \caption{Training Loss Curves}
    \label{fig:losscurve}
\end{figure}

\begin{figure}[!ht]
    \centering
    \includegraphics[width=0.72\textwidth]{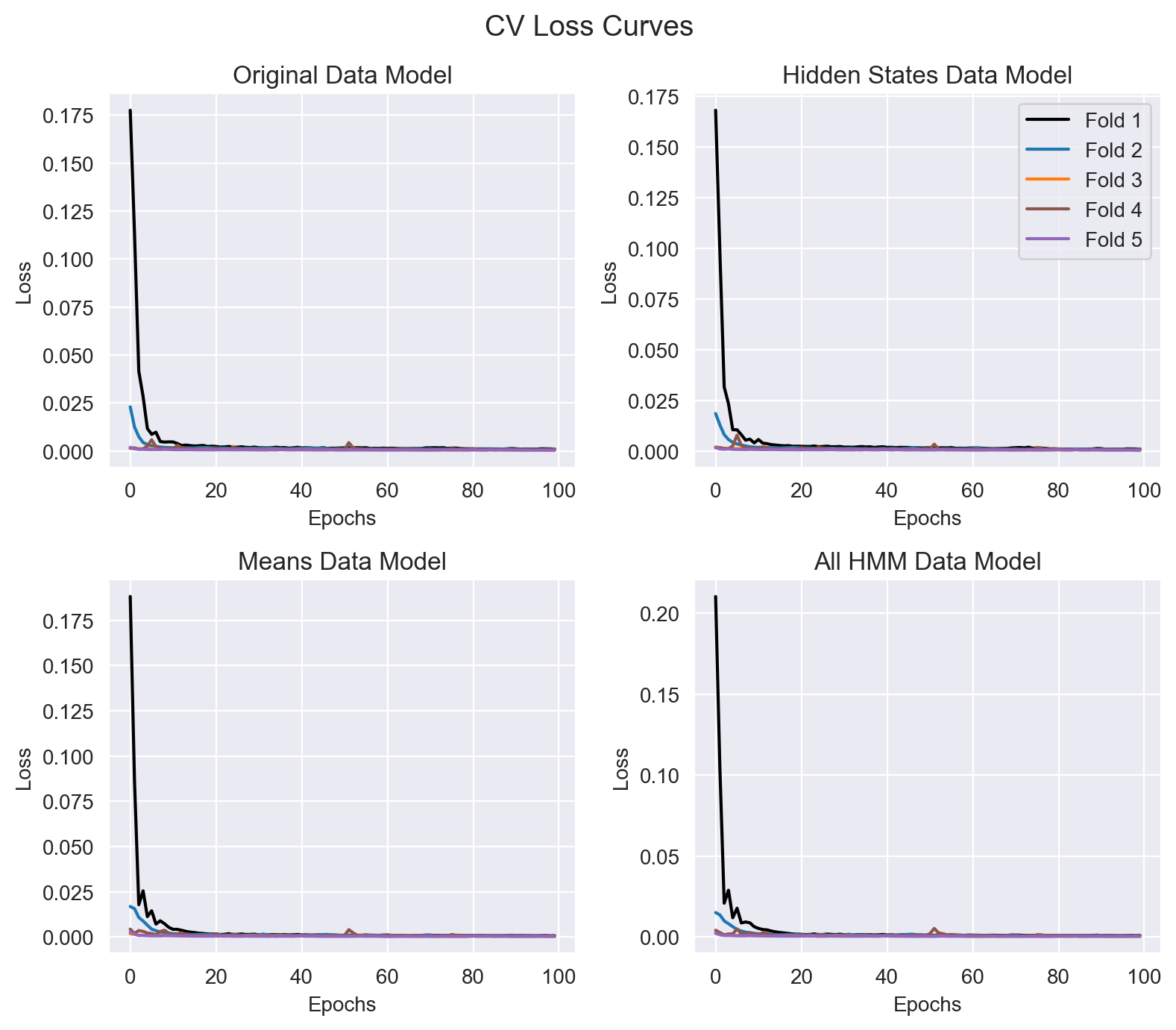}
    \caption{Training Loss Curves CV}
    \label{fig:losscurvecv}
\end{figure}

\end{document}